\RequirePackage{fix-cm}
\documentclass[twocolumn]{svjour3}          
\smartqed  
\usepackage{graphicx}
\usepackage{hyperref}       
\usepackage{url}            
\usepackage{booktabs}       
\usepackage{amsfonts}       
\usepackage{nicefrac}       
\usepackage{microtype}      
\usepackage{multirow}
\usepackage{multicol}
\usepackage{amssymb}
\usepackage{amsmath}
\usepackage{booktabs}
\usepackage{makecell}
\usepackage{array}
\usepackage{subcaption}
\usepackage[table,xcdraw]{xcolor}
\usepackage[T1]{fontenc}
\usepackage{lmodern}
\usepackage{flushend}

\newcommand{\up}[1]{\textcolor{blue}{#1}}
\newcommand{\down}[1]{\textcolor{red}{#1}}

\journalname{International Journal of Computer Vision}
\begin{document}
	
	\title{Few-shot Action Recognition via Intra- and Inter-Video Information Maximization}
	
	
	\author{Huabin Liu \and 
		Weiyao Lin \and
		Tieyuan Chen \and
		Yuxi Li \and \\
		Shuyuan Li \and 
		John See 
	} 
	
	
	\institute{
		Huabin Liu \and Weiyao Lin \and Tieyuan Chen \and Yuxi Li \and Shuyuan Li \at
		Department of Electronic Engineering, Shanghai Jiao Tong University, Shanghai, China \\
		\email{\{huabinliu,wylin,lyxok1,shuyuanli\}@sjtu.edu.cn}
		\and
        Tieyuan Chen \at
		Shanghai Jiao Tong University, Shanghai, China \\
		\email{tychen0512@gmail.com}
        \and
		John See \at
		Heriot-Watt University, Malaysia \\
		\email{j.see@hw.ac.uk}
		\and
	}
	
	\date{Received: date / Accepted: date}
	
	\maketitle
	
	\begin{abstract}
		Current few-shot action recognition involves two primary sources of information for classification:(1) \emph{intra}-video information, determined by frame content within a single video clip, and (2) \emph{inter}-video information, measured by relationships (e.g., feature similarity) among videos.      However, existing methods inadequately exploit these two information sources. In terms of intra-video information, current sampling operations for input videos may omit critical action information, reducing the utilization efficiency of video data. For the inter-video information, the action misalignment among videos makes it challenging to calculate precise relationships. Moreover, how to jointly consider both inter- and intra-video information remains under-explored for few-shot action recognition. To this end, we propose a novel framework, Video Information Maximization (VIM), for few-shot video action recognition. VIM is equipped with an adaptive spatial-temporal video sampler and a spatial-temporal action alignment model to maximize intra- and inter-video information, respectively. 
		The video sampler adaptively selects important frames and amplifies critical spatial regions for each input video based on the task at hand. This preserves and emphasizes informative parts of video clips while eliminating interference at the data level.  
        The alignment model performs temporal and spatial action alignment sequentially at the feature level, leading to more precise measurements of inter-video similarity. Finally, These goals are facilitated by incorporating additional loss terms based on \textit{mutual information} measurement. Consequently, VIM 
        acts to maximize the distinctiveness of video information from limited video data. Extensive experimental results on public datasets for few-shot action recognition demonstrate the effectiveness and benefits of our framework.
		\footnote{This manuscript is an extended version of two conference papers (\emph{TA$^{2}$N}~\cite{ta2n2022} and  \emph{Sampler}~\cite{sampler2022}). We have cited them in the manuscript and extended them substantially but not limited to the following aspects: (1) We provide a novel insight from the perspective of video information maximization to address the few-shot action recognition problem. (2) Based on mutual information measurement, we newly introduce auxiliary loss terms to facilitate the maximization of intra- and inter-video information, respectively. (3) More ablation studies on both the video sampler and action alignment model.  (4) More analysis about the interaction between video sampler and action alignment (5) More experiments are appended to demonstrate the generality and effectiveness of VIM, including upper bound analysis, limit test, many-way few-shot experiment, results of domain generalization, and more visualizations.
		}
	\end{abstract}

	\begin{figure*}[htbp]
		\centering
		\begin{minipage}{1.0\linewidth}
			\centering
			\includegraphics[width=0.9\linewidth]{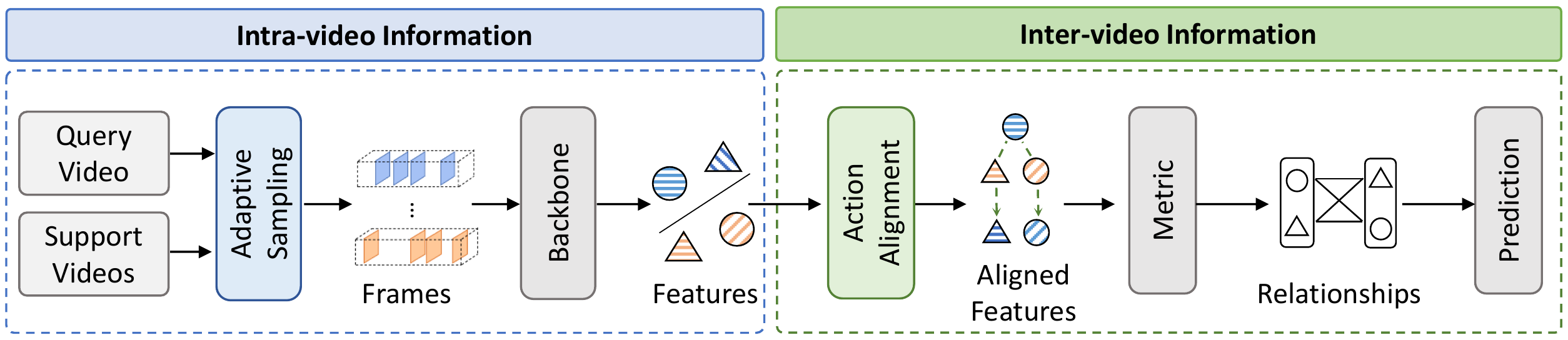}
			\caption{Overall pipeline of VIM.  The few-shot action recognition involves two primary sources of information: (1) Intra-video information and (2) Inter-video information.  Our proposed VIM newly introduces an adaptive sampling and action alignment, which seeks to maximize the inter- and intra-video information, respectively.}
			\label{fig:intro}
		\end{minipage}
		\begin{minipage}{0.49\linewidth}
			\centering
		\includegraphics[width=0.9\linewidth]{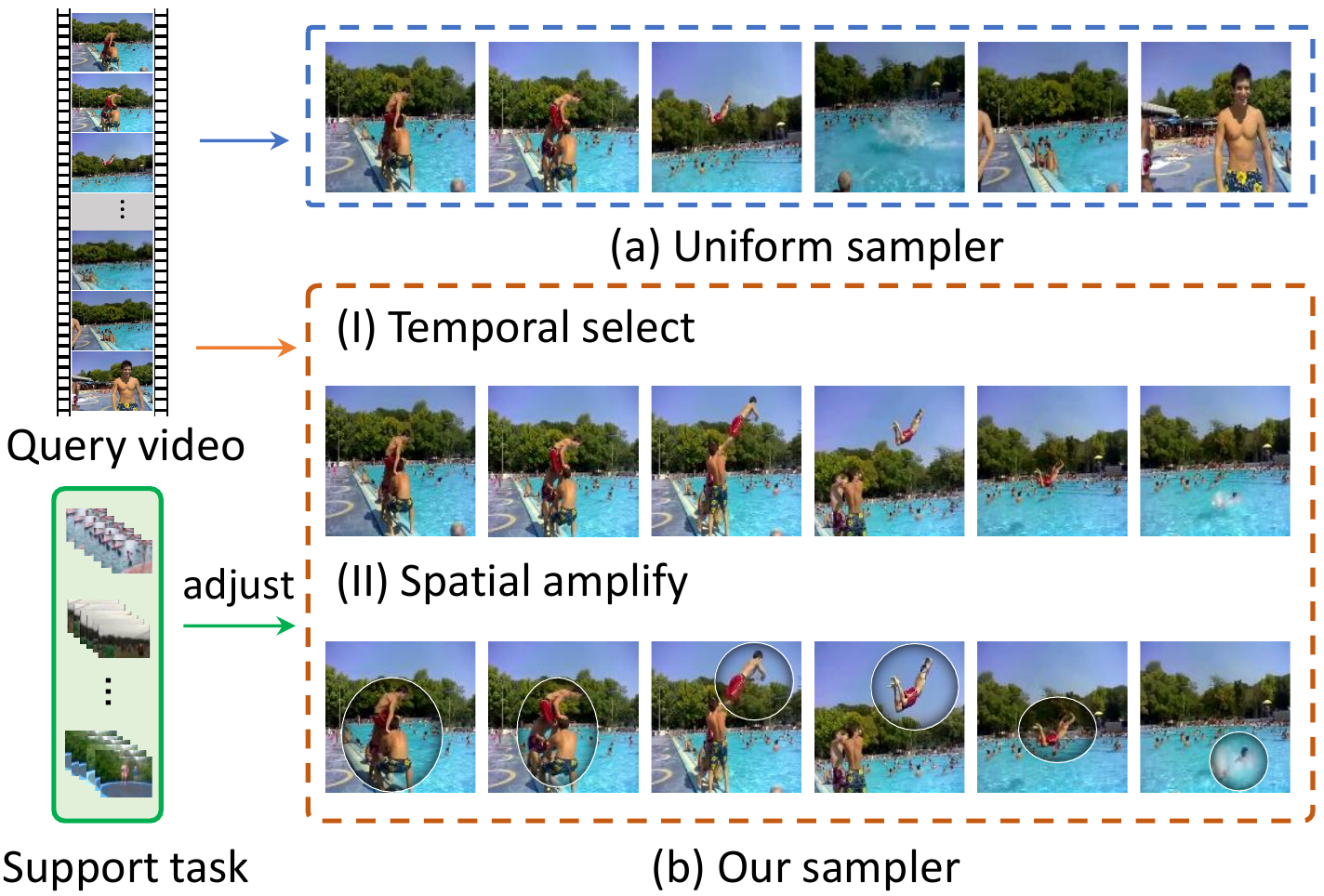}
			\caption{(a) Uniform sampler may overlook frames containing key actions. Critical regions involving the actors and objects may be too small to be properly recognized. (b) Our sampler is able to (I) select frames from an entire video that contribute most to recognition, (II) amplify discriminative regions in each frame. This sampling strategy is also dynamically adjusted according to the episode task at hand.}
			\label{fig:intro_sampler}
		\end{minipage}
		\hspace{4pt}
		\begin{minipage}{0.49\linewidth}
			\centering
			\includegraphics[width=0.95\linewidth]{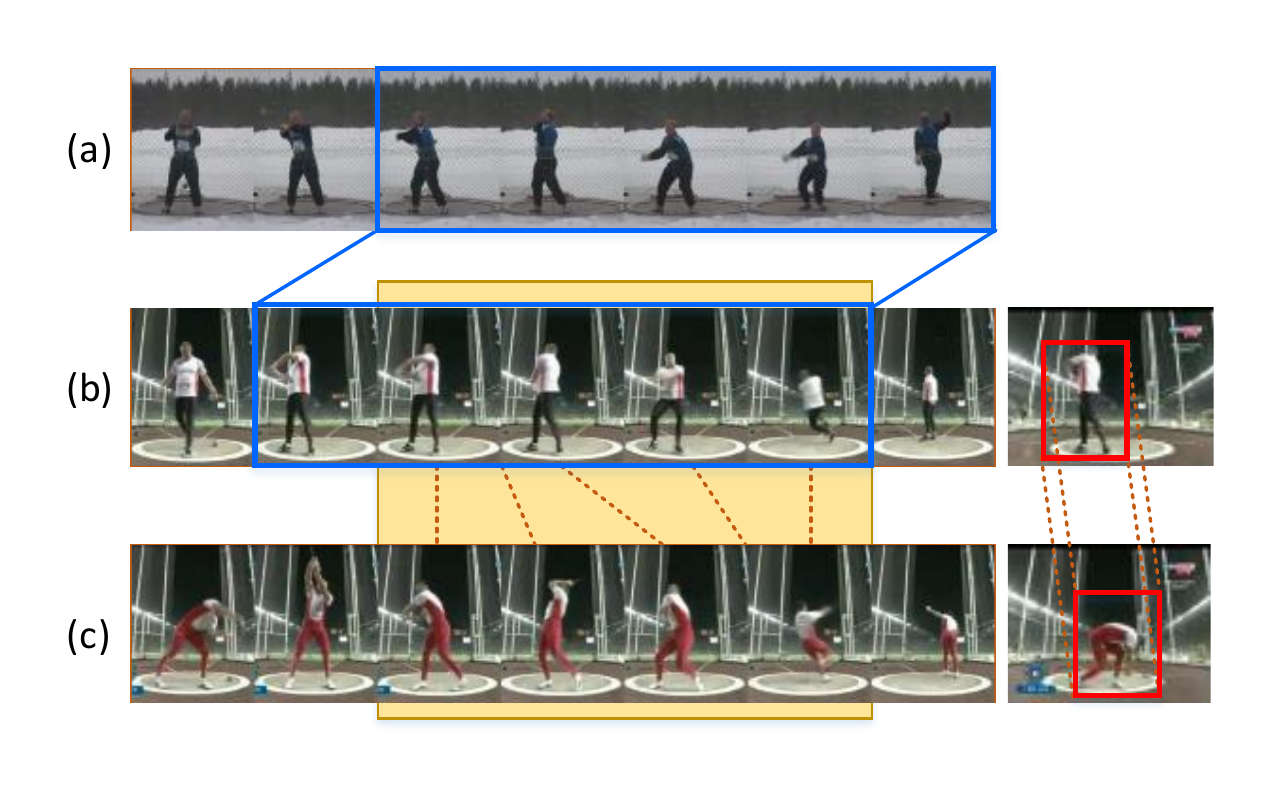}
			\caption{Illustration of action misalignment. (a)\&(b): \textit{action duration misalignment}. The action duration  is highlighted with blue rectangles. (b)\&(c): \textit{action evolution misalignment} in temporal (left) and spatial (right) aspects. The red dashed connected lines indicate pairs of temporal or spatial positions are consistent in action content. The action category of these presented videos is \textit{`Hammer throw'}.}
			\label{fig:intro_align}
		\end{minipage}
	\end{figure*}
	
	\section{Introduction}
	Recent years have witnessed spectacular developments in video action recognition~\cite{hieve2020,keypoint_comp2020,yufeng}. Most current approaches for action recognition employ deep learning models~\cite{c3d2015,I3d2017,TSN2016,survery_action}, which are expected to achieve higher performance but require large-scale labeled video data for training. Due to the expensive cost of collecting and annotating video data, sometimes very few video data samples are available in real-world applications. Consequently, the \emph{few-shot} video recognition task, which aims to learn a robust action classifier with very few training samples, has attracted much attention. 
	
	The majority of existing approaches~\cite{cmn-j2020,trx2021,tarn2019} for few-shot action recognition are built upon the metric learning paradigm.  Specifically, it first performs uniform frame sampling over each video to generate input frames. Then, a feature extractor (e.g., ResNet) is leveraged to obtain the feature representation of each video individually.   Subsequently, relationships among videos are measured using specific metrics (e.g., Euclidean or cosine distance) at the feature level. The categories of query videos can then be inferred based on their relationships with support videos. Under this pipeline, we establish that the information contributing to the final action recognition can be summarized into two distinct types (as shown in  \autoref{fig:intro}): (1) Intra-video information, which is determined by input frame contents within a single video clip and then encoded into the feature representation, and (2) Inter-video information, which is measured by the relationships among videos and directly impacts prediction.
	
	This naturally raises the question: \textit{How can we obtain more valuable intra- and inter-video information for recognition from a limited amount of data?} Driven by this inquiry, we further observe that the existing approaches have not fully exploited these two types of video information in the following aspects:
	
	\begin{itemize}
		\item For intra-video information, most approaches primarily focus on feature representation learning~\cite{cmn-j2020,ARN2020,HCL}, while little attention is paid to the processing of input videos. As illustrated in \autoref{fig:intro_sampler}(a), they typically employ uniform frame sampling along both spatial and temporal dimensions over each video to produce its corresponding frame input. This may omit critical action content and is vulnerable to some redundancy and interference from irrelevant parts in the video clip, thereby reducing the utilization efficiency of intra-video information. 
		\item For inter-video information, since different action instances show distinct spatial and temporal distributions, severe action misalignment issues exist between videos (illustrated in \autoref{fig:intro_align}). This makes it difficult to measure the relationship among videos accurately. Although some approaches~\cite{otam2020,tarn2019,trx2021} have tried to solve this issue, they lack specific analysis or guidance on how to cope with different types of misalignment. Additionally, there has been little exploration into spatial action alignment. 
		\item Additionally, to our knowledge, no research in this field has jointly considered intra- and inter-video information simultaneously, despite the fact that they are complementary and may benefit from each other. 
	\end{itemize} 
	
	To overcome these limitations, we devise a novel framework called Video Information Maximization (VIM) for few-shot action recognition. As illustrated in \autoref{fig:intro}, VIM introduces two innovative steps to the conventional recognition pipeline: video adaptive sampling and action alignment. The former targets intra-video information maximization at the data level, while the latter focuses on inter-video information maximization at the feature level. In this framework, we implement adaptive sampling by a task-specific spatial-temporal video sampler, which improves the utilization efficiency of videos via temporal selection and spatial amplification over frames. Meanwhile, action alignment is achieved through a spatial-temporal action alignment model. It learns to address misalignment issues in both spatial and temporal dimensions, resulting in more accurate measurements of the relationships among videos. In this way, the video sampler and action alignment steps are complementary to each other in VIM to maximize the overall video information.
	
    Specifically, in the VIM, input videos are first processed by the video sampler, which consists of a temporal selector (TS) and a spatial amplifier (SA).          The TS selects $T$ frames from the entire video that contribute most to the recognition, while SA learns to amplify critical action regions for each selected frame.          Moreover, the above sampling strategy can be dynamically adjusted conditioned on the task at hand.          Then, the action alignment is applied between paired video features, sequentially performing a temporal and spatial alignment.           The temporal coordination (TC) module temporally aligns the paired videos into a consistent action duration and evolution.           The spatial coordination (SC) module further ensures the action-specific regions (e.g., the actor or interacting objects) of paired frames are spatially consistent in their positions.          Finally, based on these well-aligned features, we can calculate accurate relationships between videos.      Furthermore, inspired by the application~\cite{mi_representation,vsd} of mutual information (MI) in feature representation learning, we further introduce new auxiliary loss terms based on MI measurement to facilitate the optimization of VIM.         They play their roles by (1) encouraging the video sampler to preserve more class-related information from the entire video and (2) enforcing the alignment process to provide additional class-specific information than those without performing an alignment.  Under such guidance, our VIM can be optimized toward intra- and inter-video information maximization.

	In summary, our main contributions are as follows:
	\begin{itemize}
		\item We analyze the few-shot action recognition problem from the perspective of intra- and inter-video information. Based on this, we propose a novel framework VIM that jointly maximizes these two types of video information.
		\item \sloppy Within VIM, we devise a task-specific spatial-temporal video sampler for few-shot action recognition, which significantly improves utilization efficiency in video data by adaptive spatial-temporal sampling over frames.
		\item We resolve the misalignment issues from both temporal and spatial aspects for few-shot action recognition by introducing a spatial-temporal action alignment model that addresses distinctive action misalignment sequentially.
		\item Under the criterion of video information maximization, we design two auxiliary loss terms based on the \textit{Mutual Information} (MI) measurement to facilitate the overall optimization of VIM.
	\end{itemize}
	
	\section{Related works}

    \subsection{Few-shot image classification}
    A primary challenge faced in few-shot learning (FSL) is the insufficiency of data in novel classes. The direct approach to address this is to enlarge the sample size by data augmentation. Some approaches~\cite{fsl-augment1,fsl-augment2} were proposed to generate unseen data with labels to enrich the feature spaces of novel classes. Autoaugment~\cite{autoaug2018} further learns the augmentation policy automatically to improve the generalization on various few-shot datasets. Besides, learning metrics to compare the seen and novel classes is another popular way of handling FSL. Matching network~\cite{matching2016} is an end-to-end trainable kNN model using cosine as the metric, with an attention mechanism over a learned embedding of the labeled samples to predict the categories of the unlabeled data. The Prototypical Network~\cite{prototypical2017} uses a feed-forward neural network to embed the task examples and perform nearest neighbor classification with the class centroids. Relation Net~\cite{relationNet2018} proposed a novel network that concatenates the feature maps of two images and proceeds to employ the concatenated features to a relation net to learn the similarity between features. While these methods perform well on image recognition tasks, it is less optimal
    to transfer them directly to action recognition.
	
	\subsection{Few-shot action recognition} 
	From our perspective of video information, the majority of current studies on few-shot action recognition can be interpreted as solutions to improve either inter-video or intra-video information.    
 
	Alignment-based methods are commonly used to address temporal action misalignment,  enabling more accurate measurements of inter-video information.   TARN~\cite{tarn2019} devises an attentive relation network to perform implicit temporal alignment at the video segment level, while OTAM~\cite{otam2020} explicitly aligns video sequences with a variant of the Dynamic Time Warping (DTW) algorithm.    TRX~\cite{trx2021} represents video by exhaustive pairs and triplets of frames for further alignment and matching, which is computationally expensive but leads to significant improvement.    Recently, some methods further perform alignment in the spatial dimension for videos.    ATA~\cite{ATA2022} conducts appearance alignment using frame-level feature matching to calculate appearance similarity scores. MetaUVFS~\cite{metaUVFS2021} obtains the action-specific spatial representation by aligning the frame-wise appearance features produced by 2D-CNNs to the video-wise motion features produced by 3D-CNNs.    Some methods focus on metric design to facilitate the measurement of inter-video information.    HyRSM~\cite{HyRSM2022} reformulates the distance calculation between videos as a set matching problem and adopts the Mean Hausdorff Metric for measurement.   However, the issue of action misalignment among videos has not been thoroughly analyzed from both temporal and spatial perspectives. Furthermore, most of these alignment-based methods lack specific guidance on how to align actions accurately. Additionally, there has been little exploration into spatial action alignment, despite its potential benefits.
	
	Another line of methods can be regarded as approaches that aim to maximize intra-video information. The early work CMN~\cite{compound2018} achieves this objective using a memory network to store matrix representations of videos.    ARN~\cite{ARN2020} utilizes a self-supervised training strategy to improve the robustness of video representation.    All the above methods focus on feature representation enhancement, but little attention is paid to the input video data.    AmeFu~\cite{AmeFu2020} alleviates the extreme data-scarcity problem by introducing an additional depth modality for videos. However, to our knowledge, no attempt has been made to improve the utilization of intra-video information from the root -- the input videos.
	
	More importantly, none of the existing methods can jointly maximize both inter- and intra-video information, both of which are deemed critical for improving few-shot action recognition.

	\subsection{Frame sampling in action recognition}
	Recently, some methods have attempted frame selection on input videos to improve inference efficiency.  Since the selection operation is discrete, many methods utilize reinforcement learning~\cite{adaframe2018,adafocus2021,marl2019} or the Gumbel trick~\cite{arnet2020} to address this issue.  Adaframe~\cite{adaframe2018} adaptively selects a small number of frames for each video for recognition, which is trained with the policy gradient method.  MARL~\cite{marl2019} adopted multi-agent reinforcement learning to select multiple frames in parallel fashion.  FrameExit~\cite{frameexit2021} introduced the early stop strategy for frame sampling whereby frame selection stops when a pre-defined criterion is satisfied.  Most of these works only focus on temporal selection.  A recent work AdaFocus~\cite{adafocus2021} leverages reinforcement learning to locate a specific sub-region for each frame and passes it to classification.

    While most of these methods aim to reduce the number of frames to improve inference speed, we focus on selecting a fixed number but more informative frames to improve the utilization efficiency of video (which can be regarded as a fixed-size subset selection problem). It is also worth noting that the success of these methods depends on having sufficient training data, whereas only a limited number of training videos are available in few-shot recognition. Additionally, the sampling operation on the spatial dimension, which plays a vital role in understanding video content, has not been thoroughly explored in  conventional and few-shot action recognition. 

	\begin{figure*}[th]
		\centering
	 \includegraphics[width=0.9\linewidth]{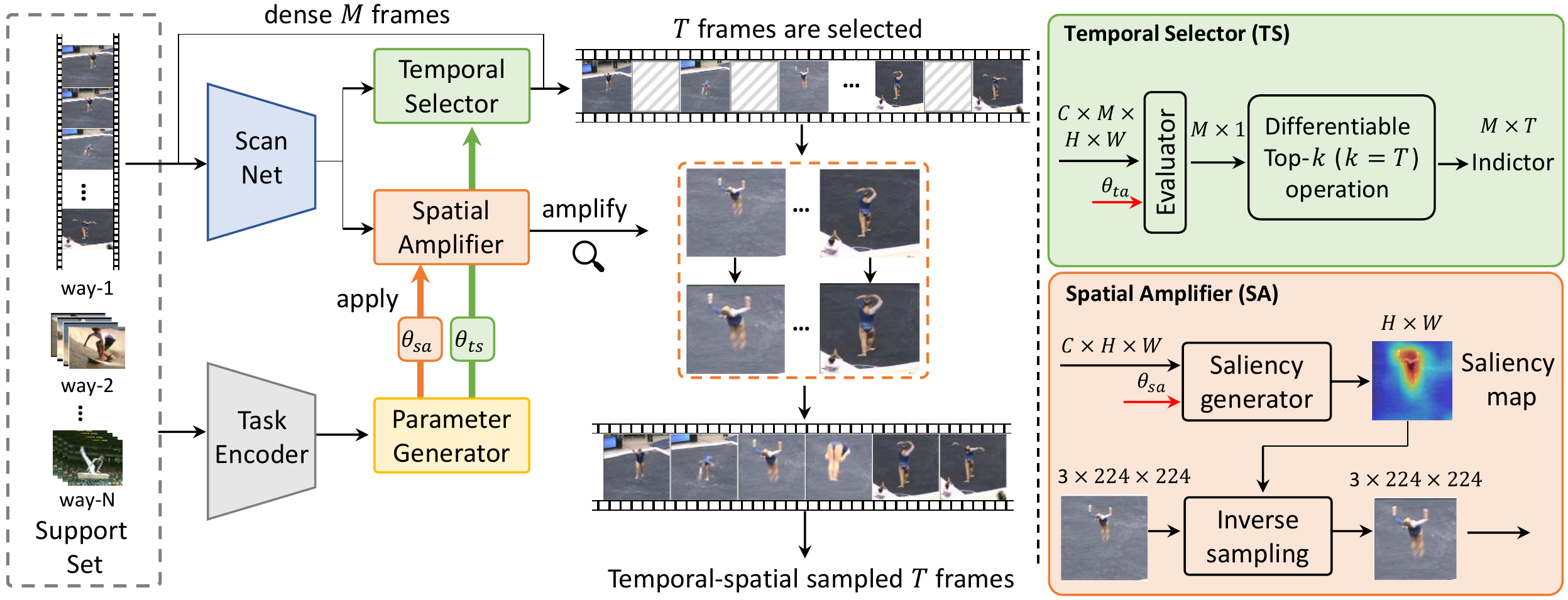}
		\caption{The overall framework of video sampler in VIM. Specifically, videos are first sent to the Scan Network with dense frame sampling to get their global perception. Based on it, TS (temporal selector) evaluates frames and selects $T$ frames that contribute most to recognition. Then selected frames are processed by SA (spatial amplifier), which amplifies discriminative sub-regions under the guidance of saliency maps. An encoder aggregates the task information and generates task-specific parameters for TS and SA to conduct a dynamic sampling according to the episode task at hand.}
        \label{fig:frameowrk_sampelr}
	\end{figure*}

	\section{Methods}
	\subsection{Problem formulation} Following standard episode training in few-shot action recognition, the dataset is divided into three distinct parts: training set $\mathcal{C}_{train}$, validation set $\mathcal{C}_{val}$, and test set $\mathcal{C}_{test}$. The training set contains sufficient labeled data for each class while there exist only a few labeled samples in the test set. The validation set is only used for model evaluation during training. Moreover, there are no overlapping categories between these three sets. Generally, \emph{few-shot} action recognition aims to train a classification network that can generalize well to novel classes in the test set. Consider an $N$-way $K$-shot episode training, each episode contains a support set $\mathcal{S}$ sampled from the training set $\mathcal{C}_{train}$. It contains $N \times K$ samples from $N$ different classes where each class contains $K$ support samples. Likewise, $Q$ samples from each class are then selected to form the query
	set $\mathcal{Q}$ which contains $N \times Q$ samples. The goal is to classify $N \times Q$ query samples using only $N \times K$ support samples. 

    \subsection{Overall pipeline of VIM}
    \autoref{fig:intro} illustrates the overall pipeline of our VIM.    Specifically, support and query videos are first fed into the video sampler.    It conducts adaptive temporal-spatial frame sampling over video clips and outputs $T$ frames for each video.    Then, we produce the feature representation for each video using the feature extractor (i.e., backbone).  Based on these features, the action alignment performs alignment between query and support videos, which generates well-aligned features for video pairs.  Finally, the relationship (e.g., distance, similarity) among videos are calculated using a specific metric.
 
	\subsection{Task-adaptive spatial-temporal video sampler}
    \autoref{fig:frameowrk_sampelr} presents the architecture of our proposed video sampler in VIM. We will elaborate on each module in the following.
	\subsubsection{Video scanning}
	
	To get a global perception of video, we first \textbf{\textit{densely}} sample $M$ frames from each video. Thus, each video is represented by a \textit{dense} frame sequence $\mathbf{X_D}=\{X_1, X_2, ..., X_M\}$, where $|\mathbf{X_D}|=M$. Our sampler performs both temporal and spatial sampling among frames in $\mathbf{X}_\text{D}$ to obtain a subset $\mathbf{X}, \text{where}~|\mathbf{X}|=T<M$. Given the dense sequence, a lightweight Scan Network $f(\cdot)$ takes the frames as input and obtains corresponding frame-level features $\displaystyle f_\mathbf{X_\text{D}} = \{f(X_1), \dots,f(X_M)\} \in \mathbb{R}^{c \times M\times h\times w}$. To reduce computation, each frame is down-scaled to a smaller size ($64\times$64 by default) before feeding into $f(\cdot)$. The video-level representation is determined as the average of all the frames $g_{\mathbf{X_D}} = \frac{1}{M}\sum_{i}^{M}f(X_i) \in \mathbb{R}^{c\times h \times w}$, for the subsequent sampling procedure. 
	

	\subsubsection{Temporal selector}
	A simple way to improve the data coverage and utilization efficiency is to feed all $M$ frames into few-shot learners without any filtering. However, this is intractable since it requires enormous computations and memory, especially under episode training (each episode contains $N(Q+K)$ videos). Therefore, we aim to select $T$ informative frames that contribute the most to few-shot recognition from the set of dense frames $\mathbf{X}$, which will improve the data utilization without introducing further overhead. To this end, a temporal selector (TS) is devised to conduct this selection. 
	
	\noindent
	\textbf{Evaluation}
	Based on frame-level features, an evaluator $\Phi_\text{eval}$ predicts the importance score for each frame. Specifically, it receives frame features as input and outputs their corresponding importance scores $\mathbf{S}\in \mathbb{R}^{M\times 1}$. Meanwhile, the global information $g_\mathbf{X}$ is concatenated with each frame-level feature to provide a global perception:
	\begin{gather}
		s_i = \Phi_{\text{eval}}(\texttt{Avg}(\texttt{Cat}(f(X_i), g_\mathbf{X})) + \texttt{PE}(i)) \\
		\Phi_\text{eval} = w_2(\texttt{ReLU}(w_1(\cdot))
		\label{eq:evaluation}
	\end{gather}
	where $w_1\in \mathbb{R}^{c\times d}$, $w_2 \in \mathbb{R}^{d\times 1}$ denotes the weights of linear layers, and $\texttt{Avg}$ denotes spatial global average pooling, $\texttt{PE}$ indicates the Position Embedding~\cite{transfomer2017} of position $i$. Specifically, the value of $w_2$ is dynamically adjusted among different episode tasks, which will be further elaborated in Sec.~\ref{sec:task-ada}. Then, scores are normalized to $[0,1]$ to stabilize the training process.
	Based on the importance scores, we can pick the $T$ highest scores by a \texttt{Top-k} operation (set $k=T$), which returns the indices $\{i_1, i_2, \dots, i_T\}\in [0,M)$ of these $T$ frames. To keep the temporal order of selected frames, the indices are sorted s.t., $i_1<i_2<\dots<i_T$. Further, the indices are converted to one-hot vectors $\mathbf{I} = \{I_{i_1}, I_{i_2}, \dots, I_{i_T}\}\in \{0,1\}^{M\times T}$. This way, the selected frame subset $X'$ could be simply extracted by matrix tensor multiplication:
	\begin{gather}
		\mathbf{X} = \mathbf{I}^{\top}\mathbf{X_D}
	\end{gather}
	where $\mathbf{I}^{\top}$ is the transpose of $\mathbf{I}$. Nevertheless, the above operations pose a great challenge in that the \texttt{Top-k} and one-hot operations are \textit{non-differentiable} for end-to-end training.
	
	\noindent
	\textbf{Differentiable top-$k$ selection}
	We adopt the \emph{perturbed maximum method} introduced in ~\cite{perturb2020,differential2021} to differentiate the sampling process during training. Specifically, the above temporal selection process is equivalent to solving a linear program of:
	\begin{equation}
		\mathop{\arg\max}_{\mathbf{I}\in \mathcal{C}}\langle \mathbf{I}, \mathbf{S}\mathbf{1}^{T} \rangle
		\label{eq:linear}
	\end{equation}
	Here, input $\mathbf{S}\mathbf{1}^{T}\in \mathbb{R}^{M\times T}$ denotes repeating score $\mathbf{S}$ by $T$ times, while $\mathcal{C}$ indicates a convex constraint set containing all possible $\mathbf{I}$.
	
	Under this equivalence, the linear program of \autoref{eq:linear} can be solved by the perturbed maximum method, which performs forward and backward operations for differentiation as described below:
	
	\noindent
	\underline{Forward} This step forwards a smoothed version of \autoref{eq:linear} by calculating expectations with random perturbations on input:
	\begin{equation}
		\mathbf{I}_{\sigma}=\mathbb{E}_{Z}\left[\underset{\mathbf{I} \in \mathcal{C}}{\arg \max }\left\langle\mathbf{I}, \mathbf{S 1}^{T}+\sigma \mathbf{Z}\right\rangle\right]
	\end{equation}
	where $\sigma$ is a temperature parameter and $\mathbf{Z}$ is a random noise sampled from the uniform Gaussian distribution. In practice, we conduct the \texttt{Top-k} ($k=T$ in our case) algorithm with perturbed importance scores for $n$ (i.e., $n=500$ in our implementation) times and compute their expectations.
	
	\noindent
	\underline{Backward} Following \cite{perturb2020}, the Jacobian of the above forward pass can be calculated as:
	\begin{equation}
		J_{\mathbf{s}} \mathbf{I}=\mathbb{E}_{Z}\left[\underset{\mathbf{I} \in \mathcal{C}}{\arg \max }\left\langle\mathbf{I}, \mathbf{S} \mathbf{1}^{\mathbf{T}}+\sigma \mathbf{Z}\right\rangle \mathbf{Z}^{T} / \sigma\right]
	\end{equation}
	Based on this, we can back-propagate the gradients through the \texttt{Top-k} ($k=T$) operation.
	
	During inference, we leverage hard \texttt{Top-k} to boost computational efficiency. However, applying hard \texttt{Top-k} during evaluation also creates inconsistencies between training and testing. To this end, we linearly decay $\sigma$ to zero during training. When $\sigma=0$, the differentiable \texttt{Top-k} is identical with the hard \texttt{Top-k}.

	\subsubsection{Spatial amplifier}
	Critical actions tend to occur in the selected regions of the frame, such as around the actors or objects. However, current approaches in few-shot action recognition treat these regions as equal to other areas (with non-critical actions) during data processing, thus reducing the efficiency of data utilization. Some works address this similar issue in image recognition by locating an important sub-region and then cropping it out from the whole image~\cite{crop1,racnn2017,scapnet2021}. However, directly replacing the original image with its partial region may further impact the data utilization efficiency under the few-shot setting. Inspired by the application of attention-based non-uniform sampling~\cite{learningzoom2018,STN2015,trilinear2019}, we adopt a spatial amplifier (SA) for few-shot action recognition, which emphasizes the discriminative spatial regions while maintaining relatively complete frame information.
	
	\noindent
	\textbf{Saliency map}
	We then introduce how the informative spatial regions of each frame are estimated. Since each feature map channel in CNN models may characterize a specific recognition pattern, we could estimate areas that contribute significantly to few-shot recognition by feature map aggregation. From the Scan Network $f(\cdot)$, we resort to using its output feature maps $f(X)\in \mathbb{R}^{c\times h \times w}$ to generate spatial saliency maps. However, there may exist activations in some action-irrelevant regions (e.g., background). Hence, directly aggregating all the channels will introduce some noises. Therefore, to enhance the discriminative patterns while dismissing these noises, we incorporate the self-attention mechanism~\cite{transfomer2017} on the channels of $f(X)$, such that
	\begin{equation}
		\alpha = \frac{f(X)f(X)^T}{\sqrt{h\times w}} \in \mathbb{R} ^ {c\times c},~~f(X)'= \alpha f(X) \in \mathbb{R} ^ {c\times h\times w}
	\end{equation}
	
	Then, we can calculate a saliency map $\mathbf{M_s}$ for frame $i$ by aggregating the activations of all channels of the feature map:
	\begin{equation}
		\mathbf{M_s}\in \mathbb{R}^{h\times w} = \frac{1}{c}\sum_{i}^{c} w_{s_i}\cdot f(X)'
		\label{eq:saliency}
	\end{equation}
	where $w_s\in \mathbb{R}^{c\times 1}$ are learnable parameters that aggregate all the channels. Note that $w_s$ is also dynamically adjusted according to the episode task at hand. We elaborate more details in \autoref{sec:task-ada}. Finally, saliency map $\mathbf{M}_s$ will be up-sampled to the same size of frame $X$.

	\noindent
	\textbf{Amplification} 
	Based on the saliency map, our rule for spatial sampling is that an area with a large saliency value should be given a larger probability to be sampled (i.e., in this context, we say that this area will be `\emph{amplified}' compared to other regions). We implement the above amplification process by the \emph{inverse-transform sampling} used in ~\cite{inverse_sample1986,trilinear2019,saliencysampling2022}. 

  \begin{figure}[t]
	\begin{center}
	\includegraphics[width=1.0\linewidth]{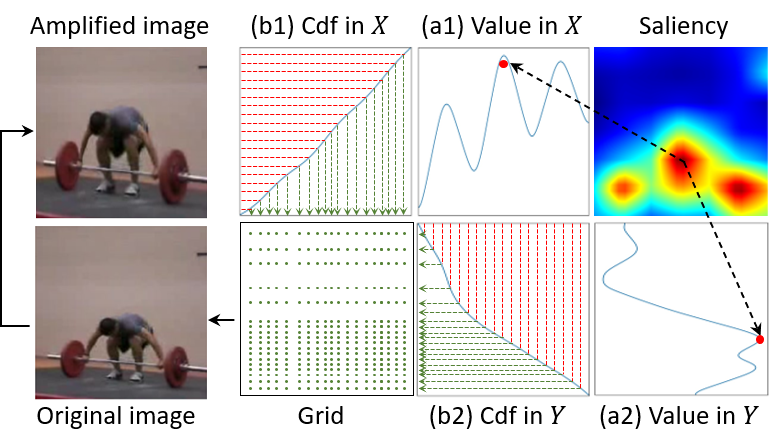}
	\end{center}
	\caption{Illustration of the 2D inverse-transform sampling.}
	\label{fig:spatial}
	\end{figure}
	
	As illustrated in \autoref{fig:spatial}, we first decompose the saliency map $\mathbf{M}_s$ into $x$ and $y$ dimensions by calculating the maximum values over axes following \cite{trilinear2019} for stabilization: 
	\begin{gather}
		M_x = \max _{1 \leq i \leq W} (\mathbf{M_s})_{i, j},~~~~M_y = \max _{1 \leq j \leq H} (\mathbf{M_s})_{i, j}
	\end{gather}
	Then, we consider the Cumulative Distribution Function (\texttt{cdf}), which is non-uniform and monotonically-increasing, to obtain their respective distributions (\autoref{fig:spatial} b1\&b2):
	\begin{gather}
		D_x = \texttt{cdf}(M_x),~~~~D_y = \texttt{cdf}(M_y)
		\label{eq:distribution}
	\end{gather}
	Therefore, the sampling function for frame $X$ under saliency map $\mathbf{M}_s$ can be calculated by the inverse function of \autoref{eq:distribution}:
	\begin{equation}
		X^{'}_{i,j} = \texttt{Func}(F, \mathbf{M_s}, i, j)=X_{{D}_{x}^{-1}(i), {D}_{y}^{-1}(j)}
		\label{eq:sample_func}
	\end{equation}
	In practice, we implement \autoref{eq:sample_func} by uniformly sampling points over the $y$ axis, and projecting the values to the $x$ axis to obtain sampling points (depicted by the green arrows shown in \autoref{fig:spatial} b1\&b2). The sampling points obtained from \autoref{fig:spatial} b1\&b2 form the 2D sampling grid. Finally, we can conduct an affine transformation above the original image based on the grid to obtain the final amplified image.
 
    The SA performs on all frames selected by TS. In this way, the amplified frame can emphasize the discriminative regions while maintaining complete information of the original one. Finally, the video sampler outputs $T$ frames $\mathbf{X}=\{X^{'}_1,\dots,X^{'}_T\}$ for each input video.
	
	\begin{figure}[t]
		\begin{center}
			\includegraphics[width=0.9\linewidth]{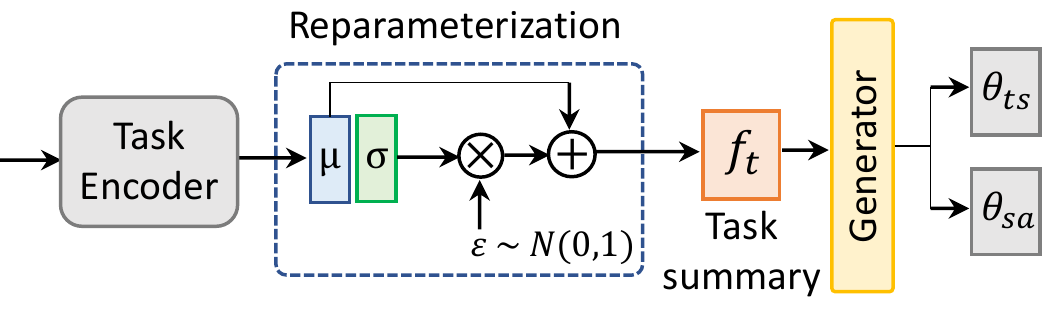}
		\end{center}
		\caption{Illustration of the task-adaptive sampling learner.}
		\label{fig:task}
	\end{figure}
	
	\subsubsection{Task-adaptive learner}
	\label{sec:task-ada}
	In general action recognition, once a sampler is well-trained, it samples each testing video with a fixed strategy and criterion~\cite{adaframe2018,arnet2020}. Nevertheless, in the few-shot episode paradigm, the query video relies on videos in the support set to perform classification. Thus, our testing videos are not independent compared to general action recognition. Therefore, fixing the sampling strategy for each video among episodes is not ideal in few-shot recognition. To this end, we adopt a task-adaptive learner for our sampler, which generates task-specific parameters for layers in TS and SA to dynamically adjust the sampling strategy according to the episode task at hand. 
	
	\sloppy Given support set $\mathcal{S}=\{(\mathbf{X}^i)_{i}^{N\times K}\}$ and its corresponding video-level features $\{(g_{\mathbf{X}^i})_{i}^{N\times K}\}$  (extracted by Scan Net) at hand, we can estimate the summary statistics of this task by parameterizing it as a conditional multivariate Gaussian distribution with a diagonal covariance~\cite{static2016,LGMNet2019}. Therefore, a task encoder $E$ is employed to estimate its summary statistics:
	\begin{equation}
		\mu, \sigma=\frac{1}{N\times K}\sum_{i=1}^{N\times K}E(\texttt{Avg}(g_{\mathbf{X}^i}))
	\end{equation}
	where $E$ consists of two linear layers, $\texttt{Avg}$ is global average pooling,  $\mu$ and $\sigma$ are 128-$dim$ mean and variance estimated by $E$. Then, we denote the probability distribution of task summary feature as:
	\begin{equation}
		\small
		p(f_t \mid \mathcal{S})=\mathcal{N}\left(\mu, \operatorname{diag}\left(\sigma^{2}\right)\right)
	\end{equation}
	We could sample a task summary feature $f_{t}$ that satisfies the above distribution with the \textit{re-parameterization} trick~\cite{vae2013}: $f_{t} \in \mathbb{R}^{128} = \mu + \sigma \varepsilon$, where $\varepsilon$ is a random variable that $\varepsilon\sim\mathcal{N}(0,1)$. In this way, each task is encoded into a fixed-length representation, which reflects more consistently across the 
	data distribution of task.
	
	Based on the summary of current task, a task-specific sampling strategy can be implemented by adjusting the parameters that influence the criterion of sampling. Specifically, we generate task-specific parameters for $w_{2}\in \mathbb{R}^{d\times 1}$ in TS (\autoref{eq:evaluation}) and $w_s\in\mathbb{R}^{c\times 1}$ in SA (\autoref{eq:saliency}) where the former decides how to evaluate the importance score of frames while the latter determines the generation of saliency maps. The above process is illustrated in \autoref{fig:task}.
	\begin{equation}
		\theta_{ts} = G_t(f_t)  ,~~\theta_{sa} = G_s(f_t) 
            \label{eq:reparam}
	\end{equation}
	where $G_t\in \mathbb{R}^{128\times d}$ and $G_s\in \mathbb{R}^{128\times c}$ denote the weights of the parameter generator. Finally, the generated parameters are normalized and filled into corresponding layers ($w_2 \leftarrow \frac{\theta_{ts}}{\Vert \theta_{ts} \Vert}$, $w_s\leftarrow \frac{\theta_{sa}}{\Vert \theta_{sa} \Vert}$).

    \begin{figure*}[t]
        \centering
        \includegraphics[width=0.9\linewidth]{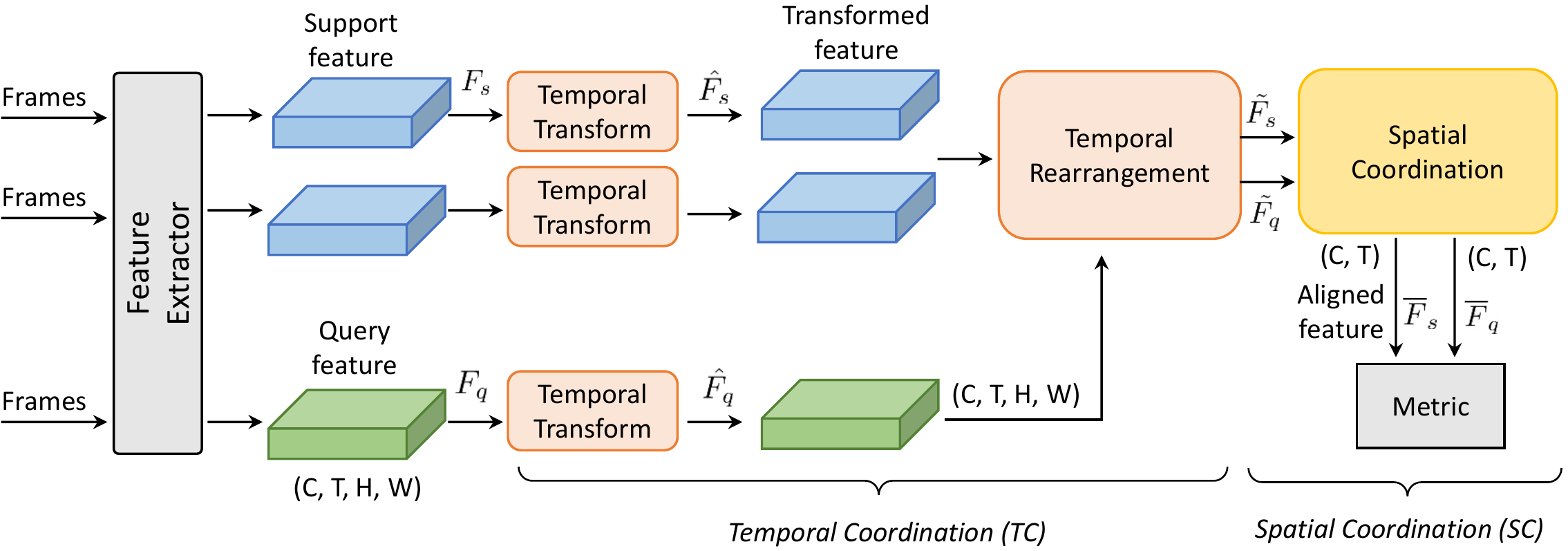}
        \caption{Overall framework of the action alignment model in VIM. Specifically, the embedded video features are first fed into the Temporal Coordination (TC) module, which addresses action duration misalignment by performing temporal transformations on each feature. Next, TC performs temporal rearrangement to align the action evolution between the support and query features. Finally, the Spatial Coordination (SC) module ensures that the support and query features are consistent in spatial position before conducting metric learning.}
        \label{fig:framework_alignment}
    \end{figure*}
	\subsection{Spatial-temporal action alignment}

    \subsubsection{Delve into the misalignment}
    We observe that the action misalignment among videos can be summarized into two distinctive types: 
    \begin{itemize}
        \item [(1)] The relative temporal location of action is usually inconsistent between videos due to different start time and duration (as shown in~\autoref{fig:intro_align}(a)\&(b)), we call the issue of location inconsistency as \textit{action duration misalignment}.
        \item [(2)] Since action often evolves in a non-linear manner, the discriminative spatial-temporal part within the action process varies from action instances (as shown in \autoref{fig:intro_align}(b)\&(c)), even though they share the same semantic category and duration. We define this internal \textit{spatial-temporal} malposition among action instances as \textit{action evolution misalignment}.
    \end{itemize}
    To cope with these two types of misalignment, we devise an action alignment model that performs temporal and spatial alignment sequentially. The overall framework of our model is presented in \autoref{fig:frameowrk_sampelr}.

    \subsubsection{Feature embedding}
    After our sampler pre-processes the raw video into $T$ frames $\mathbf{X}=\{X_1^{'},...,X_T^{'}\}$, a strong feature embedding network $F(\cdot)$ (e.g., ResNet-50) receives these frames to generate sequential features: 
    \begin{align}
        F_{\mathbf{X}}= \{F(X_1^{'}), F(X_2^{'}),\dots,F(X_T^{'})\}
    \end{align}    
    where $F_{\mathbf{X}} \in \mathbb{R}^{\rm C \times T\times H\times W}$ denotes the video representation. From this point, we will use $F_s$, $F_q$ to represent the video-level feature of the support sample and query sample, respectively.

	\subsubsection{Temporal Coordination}
	\label{sec:TTM}
    We perform the temporal alignment with a temporal coordination (TC) module. It consists of a temporal transformation and rearrangement operation, aiming to address these two types of misalignment in the temporal dimension.

	\begin{figure}[t]
		\centering
		\includegraphics[width=0.8\linewidth]{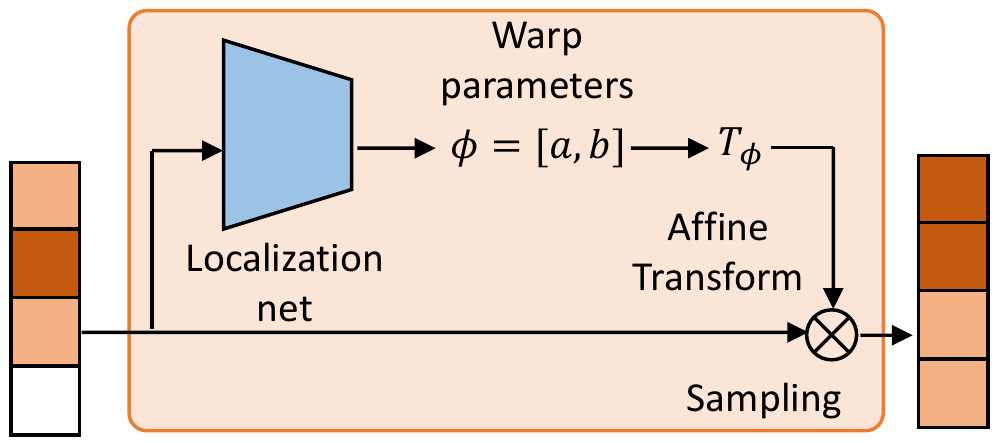}
		\caption{Illustration of temporal transformation.}
		\label{fig:T_transofmation}
	\end{figure}
    
	\noindent
	\textbf{Temporal transformation}
	To address the action duration misalignment, we aim to locate the action temporally, then the duration feature could be located and emphasized while dismissing the action-irrelevant feature. In this way, the duration misalignment could be eliminated. We achieve this objective through a temporal transformation module.
	It consists of two parts: a localization network $\mathbf{L}$ and a temporal affine transformation $\mathbf{T}$. 
	
	Specifically, given an input frame-level feature sequence $F_\mathbf{X}$, the localization network generates warping parameters $\phi=(a,b)=\mathbf{L}(F_\mathbf{X})$ firstly. Then the input feature sequence is warped by the affine transformation $\mathbf{T}_{\phi}$. 
	Succinctly, the temporal transform process is defined as:
	\begin{equation}
		\hat{F_\mathbf{X}} = \mathbf{T}_{\phi}(F_\mathbf{X}),~ \phi=\mathbf{L}(F_\mathbf{X})
	\end{equation}
	where $\hat{f_X}$ indicates the feature sequence aligned to the action duration period, $\mathbf{L}$ consists of several light trainable layers in our implementation. Since the action duration misalignment is characteristically \textit{linear} among frame sequences, the warping is represented using linear temporal interpolation. 
	
	The structure of temporal transformation is illustrated in~\autoref{fig:T_transofmation}. 
	During the episode training and testing, all the support and query features are first fed into this transformation to perform a preliminary temporal alignment, where their video features could be roughly aligned to their action periods.  In this way, the temporal transformation operation caters towards relieving the action duration misalignment problem.
	
	\noindent
	\textbf{Temporal rearrangement} To temporally align the action evolution among videos, similar motion patterns between videos should be aggregated to the same temporal location. We treat this as a global rearrangement task where the motion evolution of query video can be temporally rearranged to match the support ones. Accordingly, we model the evolution correlation $\mathbf{M}_e\in \mathbb{R}^{T\times T}$ between 
	support and query as:
	\vspace{-0.3em}
	\begin{equation}
		\mathbf{M}_e = \texttt{softmax}(\frac{(W_k\cdot G(\hat{F_s}))(W_q\cdot G(\hat{F_q}))^T}{\sqrt{dim}})
		\label{eq:tam}
	\end{equation}
	where $W_k, W_q$ indicate linear projection layer, $dim$ is the dimension of feature $G(\hat{F})$, $G$ is the global average pooling in spatial dimensions whose output tensor shape is $C\times T\times 1 \times 1$, i.e., the correlation is only calculated in the temporal dimension, $\operatorname{Softmax}$ limits the values in $\mathbf{M}_e$ to $[0,1]$.
	
\begin{figure}[t]
		\centering
		\includegraphics[width=1.0\linewidth]{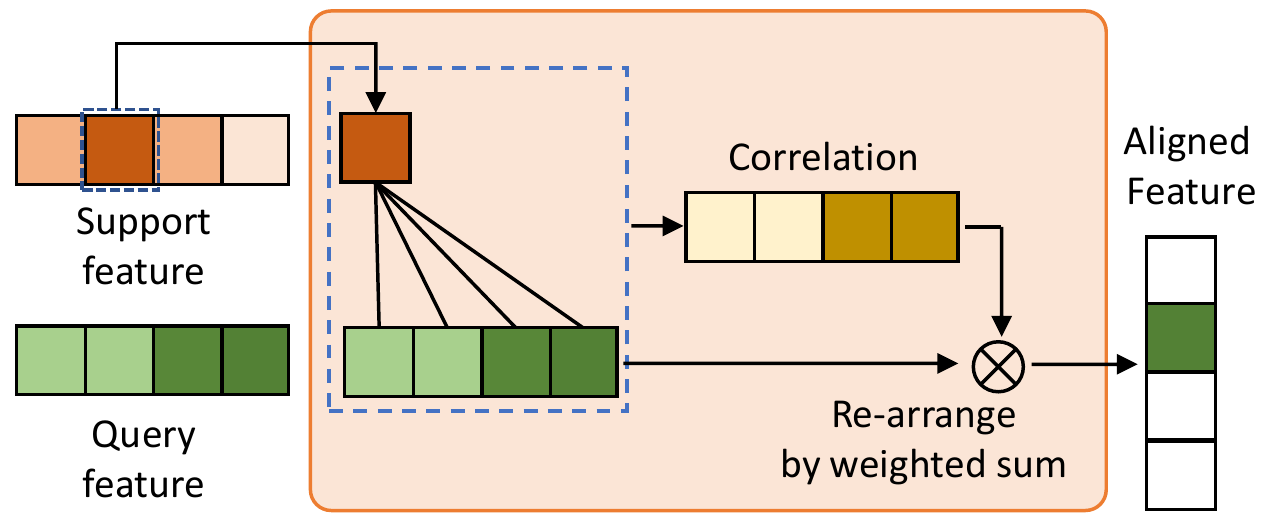}
		\caption{Illustration of temporal rearrangement.}
		\label{fig:T_coordination}
	\end{figure}

	Then, we could temporally rearrange the query feature by calculating the matrix multiplication between the normalized motion correlation matrix $\mathbf{M}_e$ and the query features:
	\begin{gather}
		\tilde{F_q}= \mathbf{M}_e\cdot (W_v\cdot G(\hat{F_q}))
	\end{gather}
	Similar to $W_k,W_q$, $W_v$ denotes a linear projection layer. In order to keep feature space consistent, this projection is also applied to support feature $\hat{F_s}$: $\tilde{F_s}= W_v\cdot G(\hat{F_s})$. This way, the same temporal location is in the consistent evolution, and thus evolution misalignment in the temporal aspect can be relieved. The illustration of TC is presented in~\autoref{fig:T_coordination}.

	\subsubsection{Spatial Coordination}
\begin{figure}
		\centering
		\includegraphics[width=0.9\linewidth]{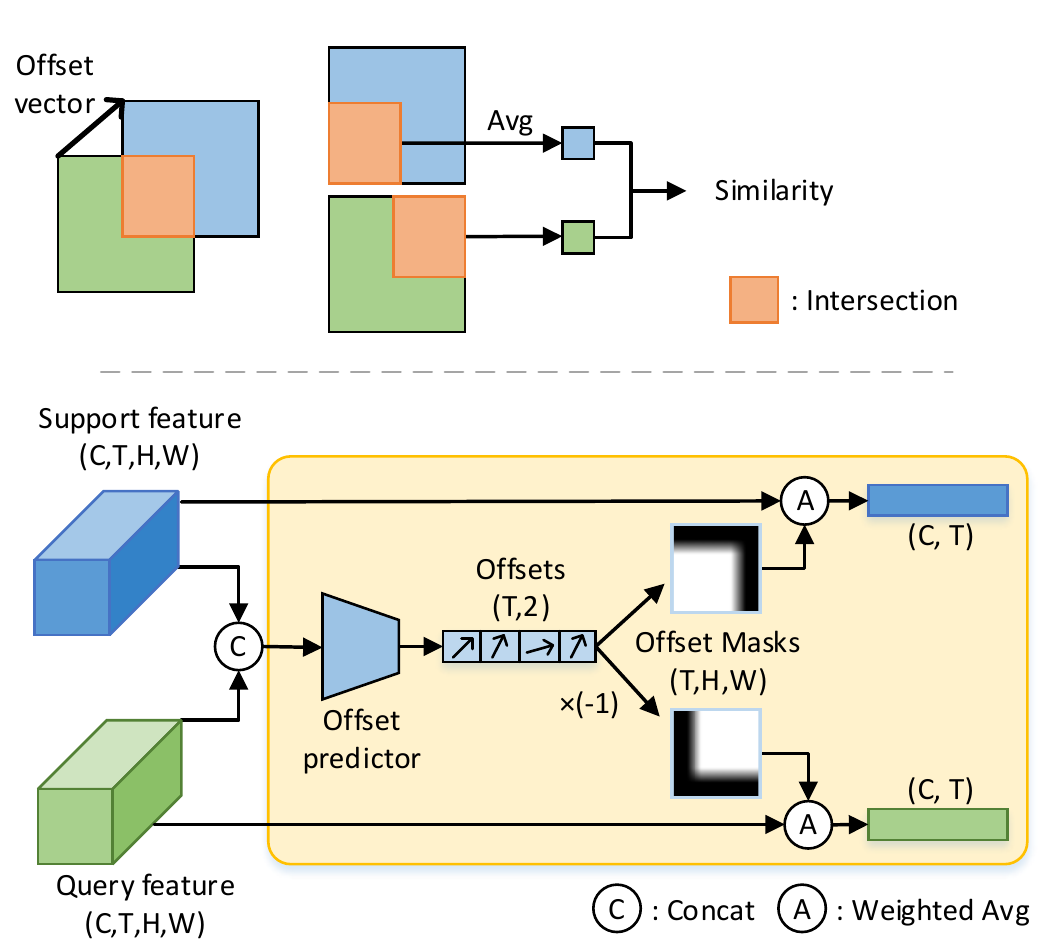}
		\caption{Spatial coordination (SC). \textit{top}: illustration of how to measure the distance under the offset. \textit{bottom}: details of offset prediction and mask generation.}
		\label{fig:SC}
	\end{figure}

	The TC ensures the action evolves with the same process over the duration period. However, the spatial variation of actor evolution, such as the positions of actors, also being critical for action recognition, which cannot be modeled by TC. Thus, we further devise the spatial coordination (SC) to reduce the spatial variations in action evolution. On the basis of temporally well-aligned features, we aim to predict a spatial offset for each paired frame and then measure their similarity in the intersection area only, as shown in the top of~\autoref{fig:SC}.
	
	Specifically, SC consists of two steps: lightweight offset prediction and offset mask generation.
	First, given the temporally well-aligned feature $\tilde{F_s}$ and $\tilde{F_q}$, they are feed into the offset predictor $S$ to predict spatial offset $O\in \mathbb{R}^{T\times 2}$ in $x$ and $y$ coordinates for all timestamps:
	\begin{gather}
		O=S(\texttt{Cat}(\tilde{F_s}, \tilde{F_q}))
        \label{eq:offset}
	\end{gather}
	where $\texttt{Cat}(\cdot)$ denotes concatenation along the channel. 
    As shown in \autoref{fig:SC}, the predicted offset indicates the relative position vector of the action-specific region between query and support frames. Then, the intersection area can be located by its corresponding spatial offset.
	
	To calculate the similarity in the intersection area in a differentiable way, for each frame, we generate an offset mask $I_o$ to calculate the average feature in the intersection on each feature. Moreover, $I_o$ is set to be smooth for differentiability, where the value of the mask is $1$ inside the intersection area and it gradually decreases to $0$ on the edge.
	Then the masks are performed over the query and support features simultaneously as weights of the weighted average:
	\begin{gather}
		\overline{F}_{s,i}=\sum_{HW}{(I_{o_i}*\tilde{F_s})} / \sum{I_{o_i}},~i=0,\dots,T\\
		\overline{F}_{q,i}=\sum_{HW}{(I_{-o_i}*\tilde{F_q})} / \sum{I_{-o_i}},~i=0,\dots,T
	\end{gather}
	
	In order to expand the exploration space of the offset predictor, we further add some perturbations on the predicted offset and use the average of corresponding features of different perturbations.
	
	Undergoing TC and SC, the action evolution misalignment among videos is dismissed in spatial-temporal aspect.
	The well-aligned paired features $\overline{F_s}$ and $\overline{F_q}$ are used in final distance measurement and classification. 
	
	\subsection{Optimization of VIM}
    \subsubsection{Classification}
    We can easily train our VIM in an end-to-end manner with standard softmax cross-entropy. Given the aligned feature of query sample $\overline{F}_q$, and the support prototype $\overline{P}_s^c$ 
	for class $c$ (the prototype is calculated as the same as ProtoNet~\cite{prototypical2017}), we can obtain the classification probability as:
	\begin{equation}
		{\rm P}(x_q \in c_i)=\frac{\exp(-d(\overline{F_q},\overline{P}_s^{c_i}))}{\sum_{c_j \in C}{\exp(-d(\overline{F_q},\overline{P}_s^{c_j}))}}
	\end{equation}
	\begin{equation}
		d(F,P)=\sum_{t=1}^{T}{1- \frac{<F_{[t]},P_{[t]}>}{\|F_{[t]}\|_2 \|P_{[t]}\|_2} },
	\end{equation}
	where $d(F,P)$ is the frame-wise cosine distance metric.
	Then the classification loss is calculated as:
	\begin{equation}
		\mathcal{L}_{\text{cls}}= -\sum_{q\in Q}{\mathbf{Y}_q \cdot \log{{\rm P}(X_q \in c_i)}},
	\end{equation}
	where $\mathbf{Y}_q \in \{0,1\}$ indicates whether $X_q \in c_i$, while $Q$ and $C$ represent the query set and its corresponding class label set, respectively.

    \subsubsection{Video information maximization}
	Furthermore, to ensure that our VIM could be well optimized towards the goal of maximizing intra-and inter-video information, we design two auxiliary loss terms based on the \textit{mutual information} measurement as additional supervision. In information theory, \textit{mutual information} (MI) measures the amount of information shared between two random variables. Formally, MI quantifies the statistical dependency of two discrete random variables:
	\begin{align}
  \mathcal{I}(V_1;V_2) = D_\text{KL}(p(V_1,V_2)||p(V_1)p(V_2))
    \label{eq:MI}
	\end{align}
    where $D_\text{KL}$ is the Kullback-Leibler (KL) divergency between joint distribution $p(V_1, V_2)$ and the product of marginals $p(V_1)p(V_2)$. \autoref{eq:MI} measures how much knowing variable $V_2$ reduces uncertainty about variable $V_1$. Particularly, $\mathcal{I}(V_1;V_2)=0$ indicates that $V_1$ and $V_2$ are independent (i.e., observing $V_2$ observing tells nothing about the prediction $V_1$). 
    
    By virtue of the above characteristic of \textit{mutual information}, some approaches~\cite{deep_infomax,tschannen2019mutual,mi_representation} learn to enhance the feature representation by optimizing the MI between input and output in neural networks. Moreover, some methods~\cite{mutual_subset1,multual_subset2} also try to solve the feature subset selection from the perspective of MI. Taking inspiration from them, we introduce additional auxiliary loss terms based on the \textit{mutual information} for our VIM to facilitate its optimization.

    \noindent
    \textbf{Intra-video information loss.}
    The video sampler aims to maximize the intra-video information at the data level by adaptive sampling over frames. Specifically, it is expected to preserve and emphasize more action-specific information while eliminating interference from the entire video. 
    From the perspective of MI, this objective can be achieved by minimizing the margin between $\mathcal{I}(\mathbf{Y}; \mathbf{X})$ and $\mathcal{I}(\mathbf{Y}; \mathbf{X_D})$:
	\begin{gather}
		\text{min}~||\mathcal{I}(\mathbf{Y}; \mathbf{X}) - \mathcal{I}(\mathbf{Y};\mathbf{X_D})||
		\label{eq:loss_intra} \\
    \text{where}~\mathbf{X}=\texttt{Sampler}(\mathbf{X_D})
	\end{gather}
	where $\mathbf{X}$ denotes the sampled frames, $\mathbf{X_D}$ denotes the dense frame sequence for scanning, $\mathbf{Y}$ denotes the label. Here $\mathcal{I}(\mathbf{Y}; \mathbf{X})$ measures the class-specific information of input videos. \autoref{eq:loss_intra} encourages the sampled frames to keep as much class-specific information as possible of the entire video clip. Therefore, we could design an auxiliary loss term for sampler to facilitate the intra-video information maximization as:
    \begin{gather}
        \mathcal{L}_{\text{intra}} = ||\mathcal{I}(\mathbf{Y}; f_{\mathbf{X}}) - \mathcal{I}(\mathbf{Y}; f_{\mathbf{X_D}})||
    \end{gather}
    where $f_\mathbf{X_D}$ and $f_\mathbf{X}$ denotes the features of $\mathbf{X_D}$ and $\mathbf{X}$, respectively.  Both of them are produced by the Scan Network $f(\cdot)$.

    \noindent
    \textbf{Inter-video information loss.}	
    The action alignment aims to provide more accurate measurements of inter-video information, which is achieved by aligning the query features to the support features. Therefore, the above objective is equivalent to maximizing the following conditional mutual information:
	\begin{gather}
		\text{max}~\mathcal{I}(\mathbf{Y};\overline{F}_q | F_q)
		\label{eq:mi_ff} \\
  \text{where}~ \overline{F}_q = \texttt{Alignment}(F_q\rightarrow F_s)
	\end{gather}
	where $F_q$ and $\overline{F}_q$ indicate the query features before/after aligning it to support feature $F_s$. This way, \autoref{eq:mi_ff} indicates the amount of class-specific information in the aligned feature $\overline{F_q}$, \textit{excluding} the information from the raw feature $F_q$. Intuitively, it measures the \textit{incremental} amount of information that results from aligning $F_q$ to support video $F_s$. Optimizing this objective will maximize the additional information we can extract from the relationships between support and query videos to improve the few-shot recognition.
	However, conditional MI is hard to be estimated in neural networks. According to the formula of conditional MI, we have:
	
	\begin{gather}
		\mathcal{I}(\mathbf{Y};\overline{F}_q | F_q) = \mathcal{I}(\mathbf{Y};\overline{F}_q) - \mathcal{I}(\overline{F}_q;F_q) + \mathcal{I}(\overline{F}_q;F_q|\mathbf{Y})
	\end{gather}
	The last term measures the class-irrelevant information shared by $\overline{F}_q$ and $F_q$, which can be omitted under the supervision of classification loss. In this way, the objective \autoref{eq:mi_ff} can be simplified to:
	\begin{gather}
		\text{max}~\mathcal{I}(\mathbf{Y};\overline{F}_q) - \mathcal{I}(\overline{F}_q;F_q)
	\end{gather}
	Therefore, we could introduce an auxiliary loss for the alignment model to facilitate the inter-video information maximization as:
	\begin{gather}
		\mathcal{L}_{\text{inter}}=-\big(\mathcal{I}(\mathbf{Y};\overline{F}_q) - \mathcal{I}(\overline{F}_q;F_q)\big)
	\end{gather}
	These MI terms can be estimated by existing MI estimators~\cite{mine,vsd}. In this paper, we employ the VSD~\cite{vsd} to estimate the MI for each term.
	
	Overall, the loss terms for training the VIM can be summarized as:
	\begin{gather}
		\mathcal{L} = \mathcal{L}_{\text{cls}} + \mu_{1}\mathcal{L}_{\text{intra}} + \mu_{2}\mathcal{L}_{\text{inter}}
  \label{eq:loss}
	\end{gather}
    where the hyper-parameters $\mu_{1}$ and $\mu_{2}$ balance different loss terms.
 
	\section{Experiments}
	\subsection{Datasets and baselines}
	\label{sec:dataset}
	
	\noindent
	\textbf{Datasets} Four popular datasets and their few-shot data splits are selected for experiments:
	\begin{itemize}
		\item UCF101~\cite{ucf101}:
		We follow the same protocol introduced in ARN~\cite{ARN2020}, where 70/10/21 classes and 9154/1421/2745 videos are included for train/val/test respectively.
		\item HMDB51~\cite{HMDB}:  We follow the protocol of ARN~\cite{ARN2020}, which takes 31/10/10 action classes with \sloppy 4280/1194/1292 videos for train/val/test.
		\item SSv2~\cite{ssv2}: For SSv2 benchmark, there are two splits proposed in CMN~\cite{compound2018} and OTAM~\cite{otam2020} denoted as $\text{SSv2}^{\dag}$ and $\text{SSv2}^{*}$ respectively.  We use the split of $\text{SSv2}^{\dag}$~\cite{compound2018} for its amount of data is closer to the standard few-shot setting, where 64/12/24 classes and 6400/1200/2400 videos are included for train/val/test.
		\item Kinetics~\cite{compound2018} contains 100 classes selected from Kinetics-400, where 64/12/24 classes are split into train/val/test set with 100 videos for each class.
	\end{itemize}
	
	\noindent
	\textbf{Competitors} The widely-used few-shot learning baseline ProtoNet~\cite{prototypical2017} is selected as the baseline of VIM. Besides, we choose recent state-of-the-art few-shot action recognition approaches, including CMN-J~\cite{cmn-j2020}, ARN~\cite{ARN2020}, TARN~\cite{tarn2019}, OTAM~\cite{otam2020}, ITANet~\cite{itanet}, TRX~\cite{trx2021}, HCL~\cite{HCL}, and HyRSM~\cite{HyRSM2022}.
	
	\subsection{Implementation details}
    We select the  ProtoNet~\cite{prototypical2017} as our baseline to conduct few-shot classification. For the video sampler, we use ShuffleNet-v2~\cite{shufflenet2018} pre-trained in ImageNet as the Scan Network $f(\cdot)$ for efficiency. In addition, we remove its first max-pooling layer to improve the quality of saliency maps. By default, we scan each video by densely sampling 20 frames at $224\times 224$ pixels (i.e., $M=20$). Moreover, each frame is downscaled to $64\times 64$ pixels before feeding into the Scan Network to reduce the computational cost. Following previous works, our sampler outputs 8 frames (i.e., $T=8$) at $224\times 224$ pixels for the subsequent few-shot learners to ensure fair comparisons. The ImageNet pre-trained ResNet-50~\cite{he2016deep} is selected as the feature extractor $F(\cdot)$ so that we could have a fair comparison with most current approaches~\cite{cmn-j2020,otam2020,trx2021,HyRSM2022}. Specifically, the feature before the last average pooling layer (stage-4) in ResNet-50 forms the frame-level input to our action alignment model. For episode training, 5-way 1-shot and 5-way 5-shot classification tasks are conducted. The whole meta-training contains 15,000 epochs, each with 200 episodes. In the testing phase, we sample 10,000 episodes in the meta-test split and report the average result. We train our model using SGD optimizer, with an initial learning rate of $1\times 10^{-2}$. The learning rate is decayed (multiplied) by 0.9 after every 5000 epochs. The temperature $\sigma$ in video sampler is set to 0.1, decayed by 0.8 after 2000 epochs. The coefficients of loss terms are $\mu_{1}=0.5,\mu_{2}=1$ in default.

 \setlength{\aboverulesep}{0pt}
    \setlength{\belowrulesep}{0pt}
	\begin{table*}[ht]
		\centering
			\begin{tabular}{c|c|c|cc|cc|cc|cc}
                \hline
				\multirow{2}{*}{Method} &
				\multirow{2}{*}{Publisher} & \multirow{2}{*}{Backbone} &
				\multicolumn{2}{c|}{HMDB51} & \multicolumn{2}{c|}{UCF101} &  \multicolumn{2}{c|}{Kinetics} & \multicolumn{2}{c}{SSv2} \\
				&  & & 1-shot & 5-shot & 1-shot & 5-shot & 1-shot & 5-shot & 1-shot & 5-shot \\ \hline
        		MAML~\cite{finn2017model} & ICML'17 & ResNet-50 &  - & - & - & - & 54.2 & 75.3 & 31.3 & 45.5\\
    			TRN++~\cite{otam2020}& ECCV'18 & ResNet-50 &  - & - & - & - & 68.4 & 82.0 & - & -\\
                Plain CMN~\cite{compound2018}& ECCV'18 & ResNet-50 &  - & - & - & - & 57.3 & 76.0& 33.4 & 46.5 \\
				ProtoGAN~\cite{protogan2019} & CVPR'19 & C3D & 34.7 & 54.0 & 57.8  & 80.2 & - & - & - & - \\
				TARN~\cite{tarn2019}& BMVC'19 & C3D&  - & - & - & - & 64.8 & 78.5 & - & -\\
				ARN~\cite{ARN2020}& ECCV'20 & C3D &  45.5 & 60.6 & 66.3 & 83.1 & 63.7 & 82.4 & - & -\\
				CMN-J~\cite{cmn-j2020}& TPAMI'20 & ResNet-50 &  - & - & - & - & 60.5 & 78.9& 36.2 & 48.8 \\
                ProtoNet*~\cite{prototypical2017}& - & ResNet-50 &  54.2 & 68.4 & 78.7 & 89.6 & 64.5 & 77.9 & 34.2 & 46.5 \\
				OTAM~\cite{otam2020}& CVPR'20 & ResNet-50 &  54.5 & 66.1 & 79.9 & 88.9 & 73.0 & 85.8 & 36.4 & 48.0\\
				AMeFu~\cite{AmeFu2020}& MM'20 & ResNet-50&  54.5 & 66.1 & 85.1 & 95.5 & - & - & - & -\\
                ITANet~\cite{itanet}& IJCAI'21 & C3D &  - & - & - & - & 73.6 & 84.3 & 39.8 & 53.7 \\
				TRX~\cite{trx2021}& CVPR'21 & ResNet-50 &  53.1 & 75.6 & 78.6 & 96.1 & 63.6 & 85.9 & 36.0 & \textbf{59.1} \\
				HyRSM~\cite{HyRSM2022}& CVPR'22 & ResNet-50&  60.3 & 76.0 & 83.9 & 94.7 & 73.7 & 86.1 & 40.6 & 56.1\\
				HCL~\cite{HCL} & CVPR'22 & ResNet-50&  59.1 & 76.3 & 82.5 & 93.9  & 73.7 & 85.8 & 38.7 & 55.4\\
				\rowcolor[HTML]{EFEFEF}
				\textbf{VIM (Ours)}& - & ResNet-50 &  \textbf{61.1} & \textbf{76.5} & \textbf{86.2} & \textbf{96.1}  & \textbf{73.8} & \textbf{87.1}& \textbf{40.8} &   54.9 \\
                \hline
			\end{tabular} 
		\caption{Few-shot action recognition results under standard 5-way k-shot setting. * refers to our re-implementation.}
		\label{tab:compare}
	\end{table*}

 	\begin{table}[t]
		\centering
		\begin{tabular}{c|ccc}
			\hline
			\multirow{2}{*}{Method} & \multicolumn{3}{c}{ActivityNet} \\ \cline{2-4} 
			& 1-shot     & 3-shot    & 5-shot         \\ \hline
			ProtoNet*                 &      66.5      &  73.4  &       80.3         \\
			OTAM*                    &       69.9     & 75.9  &         85.2       \\
			TRX*                     &       69.9     & 77.6  &        88.7        \\ 
			HyRSM*                     &     72.1     &  78.0   &    87.5    \\
			\rowcolor[HTML]{EFEFEF}
			VIM (Ours)          &      \textbf{74.7}      &  \textbf{79.3}   &         \textbf{89.6}       \\
			\hline
		\end{tabular}
		\caption{Results on ActivityNet. * refers to results from our re-implementation.}
		\label{tab:activity}
	\end{table}
	
	\subsection{Main results}
	\subsubsection{Quantitative results}
	\autoref{tab:compare} summarizes the results of state-of-the-art methods and our VIM on few-shot action recognition.    VIM outperforms various existing approaches and achieves new state-of-the-art performance in most cases.  It consistently presents significant improvements compared to the ProtoNet baseline on which it is built, across a range of benchmarks, from appearance-dominated video datasets such as HMDB, UCF, and Kinetics, to temporal-sensitive datasets like SSv2.  In fact, VIM achieves the best 1-shot and 5-shot performance on UCF, HMDB, and Kinetics datasets when compared to state-of-the-art methods.  Notably, the improvements in 1-shot performance are the most significant.  To our knowledge, VIM is the first method to improve the performance of UCF under the 1-shot setting to 86+ and HMDB under the 1-shot setting to 61+.  This observation highlights VIM's ability to learn more information for recognition when fewer samples are available.  It's also worth noting that under the 5-shot setting, VIM yields a performance of 54.9 on SSv2, which is behind TRX~\cite{trx2021} and HCL~\cite{HCL}.  Their superiority in many-shot can be attributed to the use of attention mechanisms for ensemble and feature enhancement over samples within the same category. However, this characteristic also imposes limitations on their performance in 1-shot settings.

 \subsubsection{Effectiveness on untrimmed \& long-term video}
	Current few-shot action recognition methods only conduct evaluations on short-term videos (e.g., UCF, SSv2). In this paper, we aim to benchmark our VIM  against current methods on a long-term video dataset ActivityNet~\cite{activity2015}, which has not been done before. Following the standard few-shot data split, we randomly selected 100 classes from the whole dataset. The 100 classes are then split into 64, 12, and 24 classes as the meta-training, meta-validation, and meta-testing sets, respectively. 70 samples are randomly selected for each class. ~\autoref{tab:activity} summarizes the experimental results on ActivityNet. We observe that the existing state-of-the-art methods cannot achieve satisfactory improvements on long-term videos compared to the simple baseline (ProtoNet), especially on the 1-shot and 3-shot. This may be due to the attenuation of critical actions caused by uniform sampling, which becomes more severe in long-term videos. Moreover, each video in ActivityNet may contain multiple action segments, further increasing the difficulty of metric learning. In contrast, our VIM model outperforms other methods in all cases for ActivityNet, highlighting the importance of considering both intra-video and inter-video information for achieving optimal performance in long-term action videos, particularly the intra-video information.
 
	\subsubsection{Extension to other methods}
	As illustrated in \autoref{fig:intro}, our proposed VIM does not modify the backbone architecture or metrics. As a result, it can be easily incorporated into existing few-shot action recognition methods. To further evaluate the effectiveness of VIM, we extend it into three existing approaches: OTAM~\cite{otam2020}, TRX~\cite{trx2021}, and HyRSM~\cite{HyRSM2022}. We implement these extensions based on their official code~\footnote{\url{https://github.com/tobyperrett/trx}}~\footnote{\url{https://github.com/alibaba-mmai-research/HyRSM}}. The detailed modifications of these extensions compared to vanilla VIM are as follows: 
	\begin{itemize}
		\item OTAM + VIM: It replaces the cosine distance metric with the Dynamic Time Warping (DTW) metric.
		\item TRX + VIM: It conducts an exhaustive comparison among videos based on frame tuples and triplets. 
		\item HyRSM + VIM:  Following the paradigm of HyRSM, it performs an attention mechanism over all videos within the support set and query set. It also replaces the cosine distance metric in VIM with the bidirectional Mean Hausdorff Metric (Bi-MHM).
	\end{itemize}
	The results are summarized in \autoref{tab:extension}. It can be seen that our VIM can still greatly promote these powerful methods. Although TRX~\cite{trx2021} and HyRSM~\cite{HyRSM2022} are currently considered the best methods in the 5-shot setting, incorporating VIM with them still yields stable improvements in most 5-shot cases. Furthermore, our VIM significantly improved their less-than-ideal performances in the 1-shot setting across all datasets by a significant margin.

 \begin{table}[t]
    \centering
    \resizebox{\linewidth}{!}{%
    \begin{tabular}{cc|ccc}
    \hline
    \multicolumn{2}{c|}{Extension} &
      \begin{tabular}[c]{@{}c@{}}OTAM~\cite{otam2020} \\ +VIM\end{tabular} &
      \begin{tabular}[c]{@{}c@{}}TRX~\cite{trx2021}\\ +VIM\end{tabular} &
      \begin{tabular}[c]{@{}c@{}}HyRSM~\cite{HyRSM2022}\\ +VIM\end{tabular} \\ \hline
    \multicolumn{1}{l|}{\multirow{2}{*}{HMDB}}        & 1-shot & 59.3 (\up{4.8$\uparrow$}) & 55.5 (\up{2.4$\uparrow$}) & 60.9 (\up{0.6$\uparrow$}) \\
    \multicolumn{1}{c|}{}                             & 5-shot & 74.2 (\up{8.1$\uparrow$}) & 76.3 (\up{0.7$\uparrow$}) & 75.6 (\down{0.4$\downarrow$}) \\ \hline
    \multicolumn{1}{l|}{\multirow{2}{*}{UCF101}}      & 1-shot & 83.6 (\up{3.7$\uparrow$}) & 83.7 (\up{5.3$\uparrow$}) & 85.3 (\up{1.4$\uparrow$}) \\
    \multicolumn{1}{c|}{}                             & 5-shot & 94.5 (\up{5.6$\uparrow$}) & 95.6 (\down{0.5$\downarrow$}) & 95.2 (\up{0.5$\uparrow$}) \\ \hline
    \multicolumn{1}{l|}{\multirow{2}{*}{Kinetics}}    & 1-shot & 74.3 (\up{0.7$\uparrow$}) & 68.8 (\up{5.2$\uparrow$}) & 73.9 (\up{0.2$\uparrow$}) \\
    \multicolumn{1}{c|}{}                             & 5-shot & 86.1 (\up{0.5$\uparrow$}) & 86.0 (\up{0.1$\uparrow$}) & 86.6 (\up{0.5$\uparrow$}) \\ \hline
    \multicolumn{1}{l|}{\multirow{2}{*}{SSv2}}        & 1-shot & 38.9 (\up{2.5$\uparrow$}) & 39.6 (\up{3.6$\uparrow$}) & 41.0 (\up{0.4$\uparrow$}) \\
    \multicolumn{1}{c|}{}                             & 5-shot & 51.4 (\up{3.4$\uparrow$}) & 58.0 (\down{1.1$\downarrow$}) & 56.8 (\up{0.7$\uparrow$}) \\ \hline
    \multicolumn{1}{l|}{\multirow{2}{*}{ActivityNet}} & 1-shot & 75.4 (\up{5.5$\uparrow$}) & 72.2 (\up{2.3$\uparrow$}) & 73.8 (\up{1.7$\uparrow$}) \\
    \multicolumn{1}{c|}{}                             & 5-shot & 87.7 (\up{2.5$\uparrow$}) & 89.9 (\up{1.2$\uparrow$}) & 88.1 (\up{0.6$\uparrow$}) \\ \hline
    \end{tabular}%
    }
    \caption{Extension of VIM to other methods. $\up{\uparrow}$ indicates the improvement compared to the original OTAM/TRX/HyRSM, while $\down{\downarrow}$ means decline.} 
    \label{tab:extension}
    \end{table}
 
	\subsection{Ablation study}
	\subsubsection{Effectiveness of main components of VIM}
	We first conduct an ablation study where we isolate the core components (video sampler, action alignment, $\mathcal{L}_{intra}$, and $\mathcal{L}_{inter}$ ) of VIM and quantify their effectiveness in few-shot action recognition. Results are summarized in \autoref{tab:breakdown1}. As we observe, each component yields a stable improvement over the baseline. Specifically, our experiments began by only adopting the video sampler to process all input video data. This approach demonstrated substantial improvements over the baseline on all datasets, particularly on the Kinetics and HMDB datasets. Importantly, since this setting does not apply any additional operations or modules on the feature level compared to the baseline model, it showcases that our spatial-temporal video sampler can improve the intra-video utilization efficiency at the data level, thereby boosting few-shot recognition performance.  

\setlength{\aboverulesep}{0pt}
        \setlength{\belowrulesep}{0pt}
	\begin{table*}[t]
		\small
		\centering
		\begin{tabular}{cc|cc|cc|cc|cc|cc}
            \hline
			\multirow{2}{*}{\begin{tabular}[c]{@{}c@{}}Video \\ Sampler\end{tabular}} & \multirow{2}{*}{\begin{tabular}[c]{@{}c@{}}Action \\ Alignment\end{tabular}} & \multirow{2}{*}{$\mathcal{L}_{intra}$} & \multirow{2}{*}{$\mathcal{L}_{inter}$} & \multicolumn{2}{c|}{HMDB51} & \multicolumn{2}{c|}{UCF101} & \multicolumn{2}{c|}{SSv2} & \multicolumn{2}{c}{Kinetics} \\
			&  &  &  & 1-shot & 5-shot & 1-shot & 5-shot & 1-shot & 5-shot & 1-shot & 5-shot \\ \hline
			&  & &  & 54.2 & 68.4 & 78.7 & 89.6 & 34.2 & 46.5 & 64.5 & 77.9 \\ 
			\checkmark&  & &  & 59.0 & 72.8 & 82.5 & 93.6 & 36.8 & 51.2 & 71.0 & 84.8 \\
			\checkmark&  & \checkmark &  & 59.6 & 73.2 & 83.4 & 94.3 & 37.8  & 52.8 & 72.1 & 84.9 \\
			& \checkmark &  &  & 59.7 & 73.9 & 82.8 & 95.1 & 38.6 & 53.1 & 72.8 & 85.8 \\
			& \checkmark &  & \checkmark & 60.4 & 75.6 & 85.4 & 95.7 & 39.6 & 53.7 & 73.0 & 86.2 \\
			\hline
			\checkmark & \checkmark &  &  & 59.9  & 73.5 & 83.5 & 96.0 & 38.9 & 53.8 & 73.6 & 86.2 \\
			\checkmark & \checkmark & \checkmark &  & 60.6 & 74.2 & 84.6 & 95.8 & 39.0 & 54.1 & 73.6 & 86.7 \\
			\checkmark & \checkmark &  & \checkmark  & 60.8  & 76.0 & 83.5 & 96.0 & 40.3 & 54.3 & 73.4 & 86.7  \\
			\checkmark & \checkmark & \checkmark  & \checkmark & \textbf{61.1} & \textbf{76.5} & \textbf{86.2} & \textbf{96.1} & \textbf{40.8} & \textbf{54.9} & \textbf{73.8} & \textbf{87.1} \\ 
            \hline
		\end{tabular}
		\caption{Ablation study on the effectiveness of main components of VIM. \checkmark indicates the component is adopted.}
		\label{tab:breakdown1}
	\end{table*}
	
	We then individually evaluate the action alignment model by replacing the video sampler with a normal uniform sampling strategy.  The results demonstrate that the action alignment model contributes the most to performance improvement, particularly on the SSv2 dataset (surpassing the baseline by 4.4\% accuracy). It demonstrates that the alignment among actions is critical for few-shot action recognition. The reason for this is that classification in the metric-learning paradigm is highly reliant on the accurate measurement of inter-video information, such as feature distance or similarity.  Additionally, we observe that combining the video sampler and action alignment model produces the best performance compared to a single application of them, highlighting the importance of jointly considering intra- and inter-video information maximization in few-shot action recognition. 
	
	Finally, we examined the effectiveness of our designed loss terms.   Apart from the traditional classification loss, we add the $\mathcal{L}_{intra}$ and $\mathcal{L}_{inter}$ to optimize the video sampler and action alignment during training, respectively. We observe a stable improvement in performance using any of them.  Meanwhile, when training our full VIM model with these two losses, it yields the best performance over all settings and baselines. These results prove that explicit guidance can facilitate the learning of input sampling and feature alignment, especially when only limited training data is available. Besides, it also proves that our devised losses do encourage our VIM framework to optimize towards intra- and inter-video information maximization.

        \setlength{\aboverulesep}{0pt}
        \setlength{\belowrulesep}{0pt}
	\begin{table}[t]
		\centering
		\begin{tabular}{ccc|cc|cc}
            \hline
			\multicolumn{3}{c|}{Video Sampler}       & \multicolumn{2}{c|}{Action Alignment} & \multirow{2}{*}{HMDB} & \multirow{2}{*}{SSv2} \\ \cline{1-5} 
			\textit{TS} & \textit{SA} & \textit{Ada} & \textit{TC}       & \textit{SC}       &                       &                      \\ 
			\hline
			\checkmark &   &   &   &   & 58.1 & 36.9 \\
			& \checkmark &   &   &   & 57.3 & 35.7 \\
			\checkmark & \checkmark &  &   &   & 59.1 & 37.5 \\
            \checkmark &  & \checkmark &   &  & 58.7 & 37.6  \\
             & \checkmark & \checkmark &   &   & 57.6 & 36.1 \\
			\checkmark & \checkmark & \checkmark &   &   & 59.6 & 37.8 \\
			\hline
			&   &   & \checkmark &  & 58.8 & 37.9 \\
			&   &   &   & \checkmark & 58.3 & 36.0 \\
			&   &   & \checkmark & \checkmark & 60.4 & 39.6  \\
			\hline
			\checkmark &   & \checkmark & \checkmark & \checkmark & 60.0 & 40.5 \\
			& \checkmark & \checkmark & \checkmark & \checkmark & 60.9 & 40.2 \\
			\checkmark & \checkmark & \checkmark & \checkmark &   & 60.7 & 39.1 \\
			\checkmark & \checkmark & \checkmark &   & \checkmark & 60.4 & 37.6 \\
			\checkmark & \checkmark & \checkmark & \checkmark & \checkmark & \textbf{61.1} & \textbf{40.8} \\ 
            \hline
		\end{tabular}
		\caption{Breakdown study of each module. Results are reported under 1-shot setting on HMDB and SSv2 datasets. \textit{TS}: temporal selector, \textit{SA}: spatial amplifier, \textit{Ada}: task-adaptive learner, \textit{TC}: temporal coordination, \textit{SC}: spatial coordination. \checkmark indicates the module is adopted.}
		\label{tab:breakdown2}
	\end{table}
	
	\subsubsection{Breakdown analysis on each module}
	
	In this section, we further break down all the modules of each component in VIM to analyze their impact on few-shot action recognition performance. \autoref{tab:breakdown2} summarizes the results. In terms of the video sampler, we conduct experiments with different combinations of its three modules: temporal selector (TS), spatial amplifier (SA), and task-adaptive learner (Ada). As we can see, the TS brings significant gains on HMDB and SSv2, indicating that conducting temporal selection is crucial for few-shot action recognition, which is also consistent with the conclusions derived from general action recognition~\cite{adaframe2018,adafocus2021}. Besides, we also observe that the SA works better on HMDB than SSv2 dataset. Since the videos of SSv2 are first-person scenes and the main object in frames are human hands, amplifying these hands leads to limited improvement in recognition. Further, by introducing task adaptation for TS and SA, they were able to reach their peak performances. This demonstrates that dynamic sampling strategies among episodic tasks are indeed beneficial and essential for few-shot action recognition. 
	
	As for the action alignment, we conduct experiments by first applying temporal coordination (TC) and spatial coordination (SC) separately. We observe that both of them can boost the performance of few-shot action recognition since each dimension plays an equally important role in action alignment. When TC and SC are applied in a sequential manner, the performance can be further improved, thereby supporting our notion of a spatiotemporal alignment design.
	
	Finally, we further test different combinations of modules from both the video sampler and action alignment. To our expectation, equipping all the modules yields the best performance for VIM, fully justifying the relevance of each module.

\begin{figure}[t]
\centering
	\begin{minipage}{1.0\linewidth}
		\centering
 \includegraphics[width=0.9\linewidth]{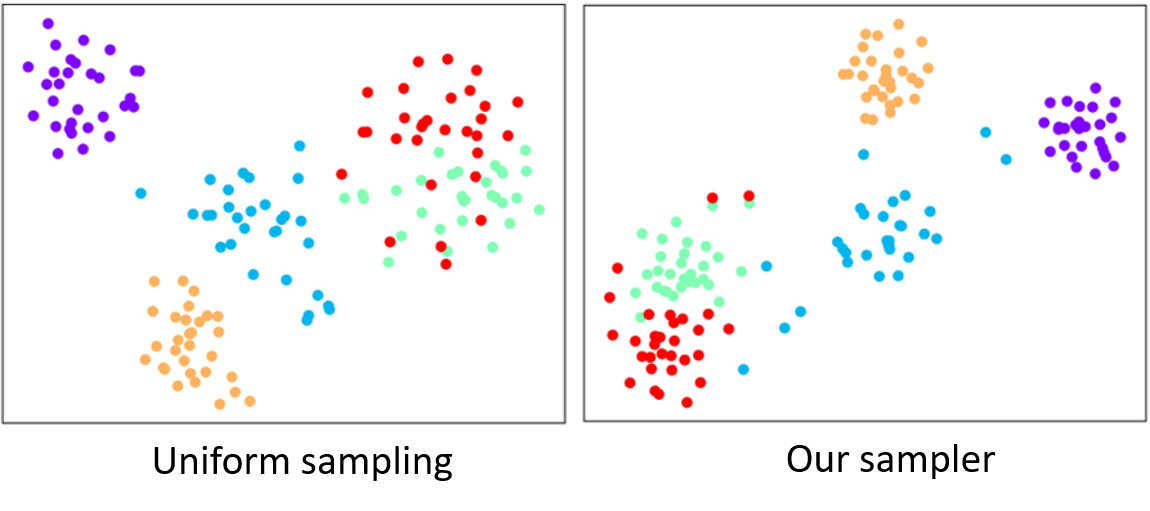}
		\caption{t-SNE feature embedding of all videos w/o and with our proposed sampler. Reported using all videos in a 5-way 30-shot episode on ActivityNet.}
		\label{fig:tsne_sampler}
  \hspace{2cm}
	\end{minipage}
	\begin{minipage}{1.0\linewidth}
		\centering
        \includegraphics[width=0.9\linewidth]{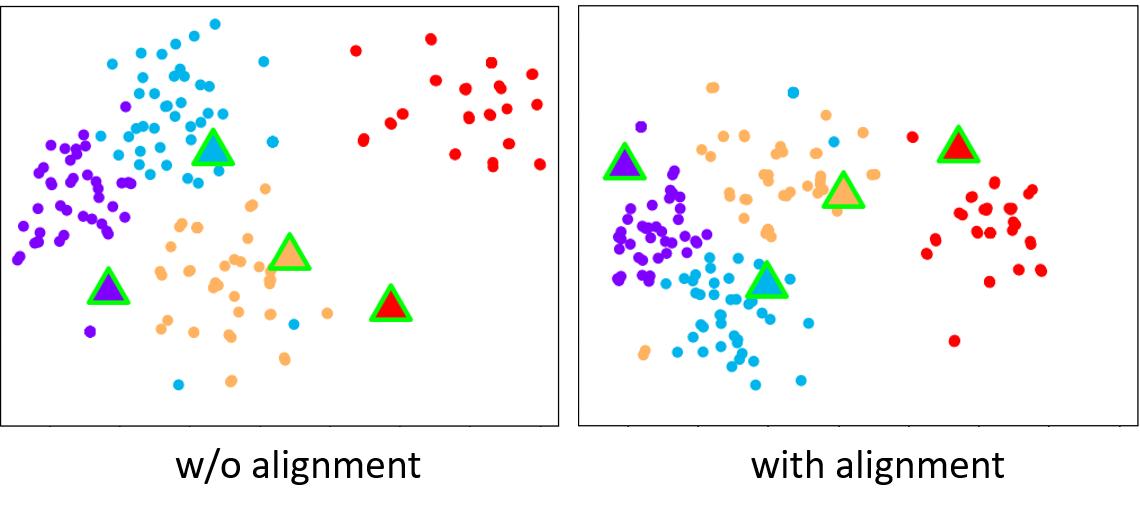}
		\caption{t-SNE feature embedding of query videos (\textit{circles}) and support prototypes (\textit{triangles}) w/o and with applying our alignment. Reported in a 4-way 20-shot episode on HMDB.}
		\label{fig:tsne}
	\end{minipage}
\end{figure}
	\subsubsection{t-SNE visualization}
    To illustrate the effect of different components in our VIM more intuitively, we visualize the feature embedding of videos under the following circumstances: (1) with and w/o adaptive sampling by our sampler, (2) with and w/o applying our action alignment. The results are presented in \autoref{fig:tsne_sampler} and \autoref{fig:tsne}, respectively. We can see that each cluster appears more concentrated after both adaptive sampling and action alignment. It demonstrates that the sampler and alignment can facilitate better feature metric learning. Moreover, for action alignment, each cluster is placed closer to its prototype after performing the alignment. It further proves that our action alignment model can well-align the query videos to support videos, obtaining a more consistent feature representation.

	\subsubsection{Analysis of video sampler}
	\noindent
	\textbf{Frequency of scanning.} To ensure that the Scan Network in the sampler can cover more frames in the videos, we set $M=20$ by default to provide a trade-off between computation and performance. To investigate the effect of video scanning frequency in our sampler, we adjust different scanning frequencies (i.e., value of M) for experiments. Results are shown in \autoref{tab:frequency}. Directly increasing the frequency cannot guarantee a consistent improvement to the baseline due to the existence of redundancy. When enlarging $M$ from 14 to 20, the performance on UCF further improves and peaks at M=20, while it brings no additional gains on the SSv2 dataset. The difference between UCF and SSv2 can be attributed to the fact that videos in SSv2 are much shorter than in UCF.  When $M=32$, there exists a large decrease in the performance of UCF and SSv2. The reason may be that learning an ideal subset selection from too many candidates is difficult using only a few training samples.
	\begin{table}[t]
		\small
		\centering
        \resizebox{\linewidth}{!}{
		\begin{tabular}{c|cccccccc}
			\hline
			M =        & 14 & 16 & 18  & 20   & 24   & 28   & 32   \\ \hline
			UCF101  & 84.8 & 85.9 & 85.9 & \textbf{86.2} & 85.5 & 84.6 & 83.9 \\ \hline
			SSv2    &  \textbf{40.8} & 40.4 & 39.8 & 39.7  & 39.4 & 38.6 & 38.3 \\ \hline
		\end{tabular}}
		\caption{Ablation of frequency of dense sampling, results are reported under 5-way 1-shot setting and $T=8$.}
		\label{tab:frequency}
	\end{table}
	\begin{table}[t]
		\small
		\centering
		\begin{tabular}{c|cccccc}
			\hline
			T =     & 4    & 6    & 8    & 10   & 12   & 14   \\ \hline
			UCF101 & 84.9 & 85.3 & 86.2 & 86.4 & 86.2 & \textbf{86.7} \\ \hline
			SSv2   & 37.8 & 39.3 & 40.8 & 40.9 & \textbf{41.2} & 41.0 \\ \hline
		\end{tabular}
		\caption{Ablation of the number of sampled frames, reported under 5-way 1-shot setting and $M=20$. Note that we only report the results under $T=8$ in \autoref{tab:compare} for \textit{fair comparisons}, although it's not necessarily the optimal setting.}
		\label{tab:number_of_T}
	\end{table}

\begin{table}[t]
	\centering
    \resizebox{\linewidth}{!}{
	\begin{tabular}{c|cc|cc}
		\hline
		\multirow{2}{*}{Scanning input frame size} & \multicolumn{2}{c|}{UCF101} & \multicolumn{2}{c}{SSv2} \\
		& 1-shot & 5-shot & 1-shot & 5-shot \\ \hline
		32$\times$32   & 85.1  & 94.8   &   39.6     &    54.1    \\ \hline
		64$\times$64 (\textbf{default})   & 86.2   & 96.1  &   40.8     & 54.9       \\ \hline
		128$\times$128 & 86.5   & 96.3   &   40.8     &  54.7      \\ \hline
		224$\times$224 & 86.5   & 96.4   &   41.1     &   54.9     \\ \hline
	\end{tabular}}
	\caption{Results of different frame size for scanning}
	\label{tab:scan_size}
\end{table}

	\noindent
	\textbf{How many frames do we need?} For fair comparisons, we follow existing methods to sample $T=8$ frames as input for few-shot learners to report our results in \autoref{tab:compare}. By virtue of our sampler, increasing the number of sampled frames is likely to introduce more informative frames for recognition instead of interference. We adjust the value of $T$ from $4$ to $14$ to explore this issue in \autoref{tab:number_of_T}. For SSv2, the performance grows with $T$ and peaks at $T=12$. However, its performance slightly drops when $T$ increases to 14. For UCF101, it is clear that increasing the number of sampled frames can consistently boost performance. On the contrary, when reducing the frame number to $4$, the SSv2 suffers a significant drop in performance while UCF101 only drops slightly. This is consistent with the observation that SSv2 is more sensitive to temporal information.

\begin{table}[t]
	\centering
	\begin{tabular}{c|c|cc}
		\hline
		\multirow{2}{*}{Scan Network} & \multirow{2}{*}{Params} & \multicolumn{2}{c}{UCF101} \\
		&                         & 1-shot       & 5-shot      \\ \hline
		ShuffleNetV2 0.5$\times$ (\textbf{default})            & \textbf{1.4M}                    & 86.2         & 96.1       \\ \hline
		ShuffleNetV2 1.5$\times$              & 3.5M                    & 86.3         & 95.9        \\ \hline
		ShuffleNetV1 2.0$\times$              & 5.4M                    & 85.9         & 95.5        \\ \hline
		MobileNetV2                   & 3.4M                    & 86.1         & 96.1        \\ \hline
		0.75 $\times$ MobileNetV1              & 2.6M                    & 85.5         & 95.4       \\ \hline
	\end{tabular}
	\caption{Different model selection for scanning net.}
	\label{tab:model_scan}
\end{table}

\begin{table}[t]
\small
\centering
\begin{tabular}{c|cc|cc}
\hline
\multirow{2}{*}{Style} & \multicolumn{2}{c|}{UCF101} & \multicolumn{2}{c}{SSv2} \\
                       & 1-shot      & 5-shot     & 1-shot      & 5-shot     \\ \hline
None                & 84.8        & 94.2       & 40.5        & 54.2      \\
Amplifying (Ours)             & \textbf{86.2}       & \textbf{96.1}       & \textbf{40.8}        & \textbf{54.9}       \\
Cropping                   & 84.7        & 93.3       & 40.6        & 53.7       \\ \hline
\end{tabular}
\caption{Results of different solutions for spatial sampling in VIM. `None' means no spatial operation is performed over video sampling.}
\label{tab:crop}
\end{table}
 
	\noindent
	\textbf{Scanning input frame size.} In our implementation, we first downsample each frame to a relatively smaller size before feeding it into the scan network to save computational costs. In this part, we study how the input size of the scan network affects the performance of our sampler. Experimental results are shown in \autoref{tab:scan_size}. Increasing the input size to $224\times 224$ only improves the overall performance by a slight margin, but it incurs significantly higher computational costs. Meanwhile, the performance drops rapidly using our smallest choice of $32\times 32$ for inputs. Therefore, we adopt $64\times 64$ as the default input size for the scan network to provide an optimum trade-off between efficiency and performance in practice.
	
	\noindent
	\textbf{Model selection of sampler.} To save computational costs and speed up the scanning, we select ShuffleNet-V2 as our scan network. In this part, we further study the impact of employing different scanning models\footnote{We adopt public models from \url{https://github.com/megvii-model/ShuffleNet-Series}}. Results are summarized in \autoref{tab:model_scan}. It can be seen that all of them share almost similar performances. Specifically, the ShuffleNet surpasses the MobileNet by a slight margin. From this, we select the ShuffleNet-V2 0.5×, which has minimal FLOPs and parameters, to build our scan network.

	\noindent
	\textbf{Spatial amplifying or cropping?} In image classification, there exists another popular solution for spatial sampling, which is to locate a sub-region and then crop it from the original image~\cite{crop1,racnn2017,adafocusv2}. However, this operation may not be ideal for few-shot learning since it reduces the data utilization efficiency in the spatial dimension. Moreover, it requires setting a fixed size for the sub-region to achieve differentiable implementation and often involves multi-stage training. In contrast, our spatial amplifier can emphasize the discriminative region while maintaining most information in the whole image. Besides, our implementation is more flexible to different degrees of amplification. To gain further insight, we substitute our SA with a crop-based solution in~\cite{adafocusv2}. In the summary of results shown in \autoref{tab:crop}, we observe that the crop-based solution only improves the 1-shot performance on the SSv2 dataset by a slight margin, while it leads to declines in other cases. This may be due to the removal of valuable recognition information from the original image during cropping. In contrast, our amplifier brings promising improvements in all settings. These findings indicate that our spatial amplification is more suitable for the few-shot circumstance, which preserves and emphasizes more intra-video information in the spatial dimension.

 \begin{table}[t]
	\centering
	\begin{tabular}{c|cc}
		\hline
		Models               & UCF101 & SSv2 \\ \hline
		ProtoNet             &   78.7     &  34.2     \\ \hline
		ProtoNet + MGSampler~\cite{mgsampler} &   80.5    &   36.8   \\ \hline
		ProtoNet + Our TS      &    82.6    & 37.6     \\ \hline
		ProtoNet + Our Sampler (with $\mathcal{L}_{\text{intra}}$)      &   \textbf{83.4}     &   \textbf{37.8}   \\
		\hline
	\end{tabular}
	\caption{Comparison with existing frame sampler in general action recognition. Reported under 1-shot.}
	\label{tab:other_sampling}
\end{table}

	\begin{figure}[t]
		\centering
		\includegraphics[width=0.6\linewidth]{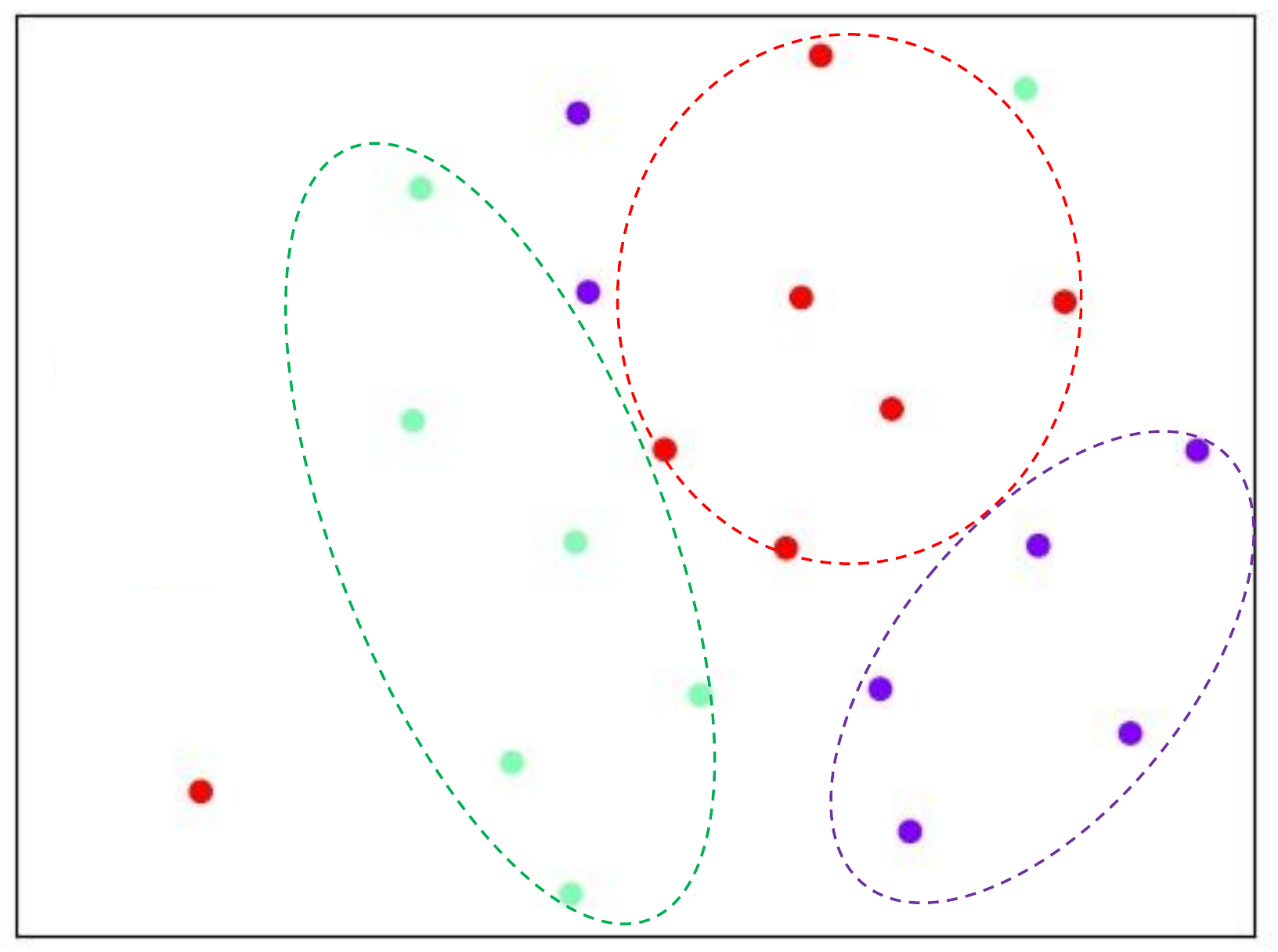}
		\caption{t-SNE visualization of task-specific weights. Different tasks are indicated by different colors.}
		\label{fig:task_tsne}
	\end{figure}

 \begin{figure*}[t]
	\centering
	\begin{minipage}[t]{1.0\textwidth}
		\centering
		\includegraphics[width=1\textwidth]{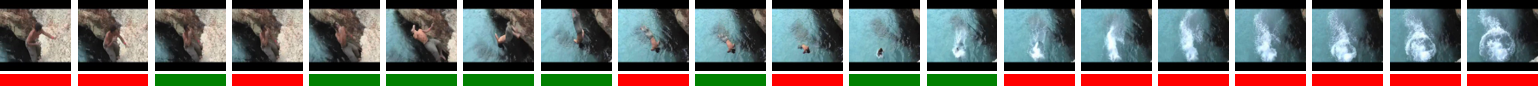}
		\subcaption{Cliffdiving.}
	\end{minipage}
	\begin{minipage}[t]{1.0\textwidth}
		\centering
		\includegraphics[width=1\textwidth]{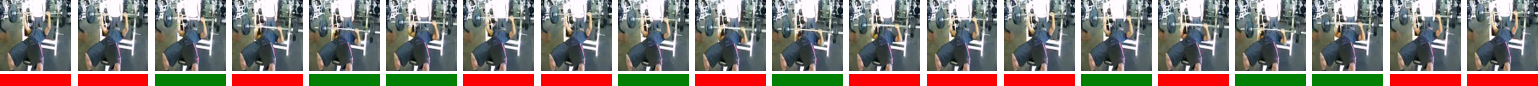}
		\subcaption{BenchPress.}
	\end{minipage}
	\caption{Visualization results of the temporal selector on two selected action classes. The green bar indicates that our sampler selects the frame.}
	\label{fig:vis_ts}
\end{figure*}
	
	\begin{figure}[t]
	\begin{center}
		\includegraphics[width=0.9\linewidth]{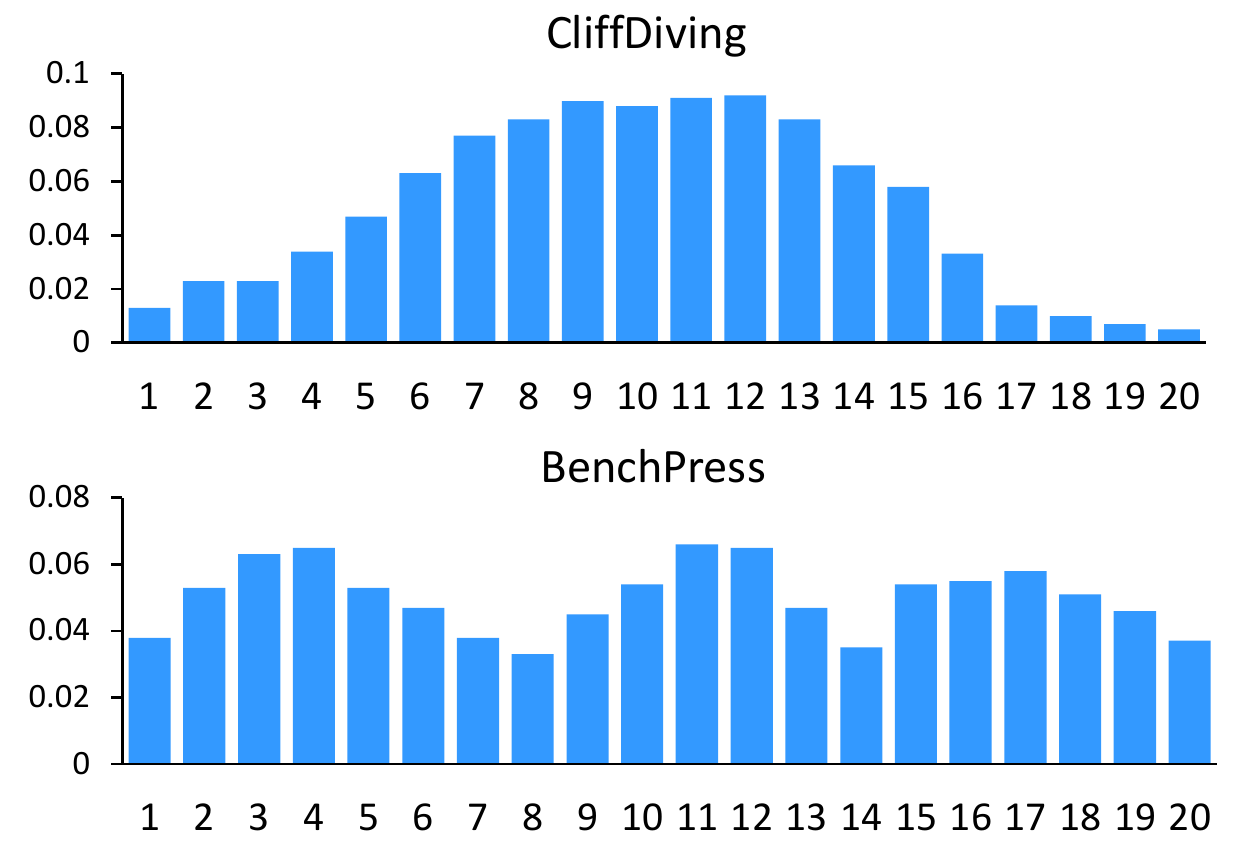}
	\end{center}
	\vspace{-1em}
	\caption{Histogram of sampled frame locations on the videos of action `\textit{CliffDiving}' and `\textit{BenchPress}'. Reported with $T=8$ and $M=20$, using 50 videos per class.}
	\vspace{-1em}
	\label{fig:distribution}
\end{figure}

	\noindent
	\textbf{Comparison to other sampling approaches.}
	 Most existing frame selection methods cannot be directly applied to address our issues (fixed-size subset selection). To this end, we apply the recently proposed MGSampler~\cite{mgsampler} to few-shot action recognition, which can be regarded as an independent pre-processing operation for video frame sampling. Specifically, it first calculates the feature difference between adjacent frames to estimate the motion process for the whole video. Then, frame selection is performed based on this estimated motion information. In this way, we feed the $T = 8$ frames selected by MGSampler based on the calculated motion difference among $M = 20$ frames into ProtoNet to report its performance. As shown in \autoref{tab:other_sampling}, the MGSampler can also improve the baselines, especially for SSv2 dataset. However, our single temporal selector (TS) still outperforms it on both datasets. Further, when our complete sampler (i.e., TS + SA) is adopted, we surpass its performance by a significant margin. The above results prove that (1) spatial sampling is necessary and beneficial, especially for appearance-dominated datasets like UCF and HMDB, (2) Directly applying current methods in general action recognition is not ideal, learning specific sampling strategies for few-shot action recognition is obviously a better solution.

	\noindent
	\textbf{Study on the task-adaptive learner.} To gain further insight into our task-adaptive learner, we visualize the generated dynamic parameters for different tasks with t-SNE. Specifically, we randomly select three different tasks and proceed to sample 7 task summary features that had undergone re-parameterization (in Eq.~\ref{eq:reparam}). We then obtain 7 task-specific parameters for $w_2\in \mathbb{R}^{d\times 1}$ by our generator. Finally, we visualize the generated weights of these three tasks with t-SNE. In results presented in \autoref{fig:task_tsne}, it can be observed that task-specific weights of the same task are situated closely in feature space, while weights of different tasks appear to have their very own distributions. It demonstrates that our task-adaptive learner is capable of generating task-specific weights of layers in our sampler network for effective sampling strategies.

	\noindent
	\textbf{Distributions of sampling.} To gain more intuitive insights into temporal frame selection, we focus on specific action classes that tend to follow a predictable distribution over time. Specifically, we select the `\textit{CliffDiving}' and `\textit{BenchPress}' classes from UCF dataset. For videos belonging to these categories, we keep count of the temporal locations of the frames selected by TS and present the distribution in \autoref{fig:distribution}. We can observe that the distribution varies between these two types of actions. For `\textit{CliffDiving},' our sampler tends to select frames from the middle of videos. This is consistent with the observation that `\textit{CliffDiving}' videos typically contain noise frames at the start (e.g., the preparation phase before diving) and the end of the video (e.g., water splash). On the other hand, for the periodic action `\textit{BenchPress},' our sampler also selects frames in a periodic manner, which is consistent with the regular pattern of this action. To further illustrate them, we visualize some results of frame selection in TS in~\autoref{fig:vis_ts}. These findings suggest that our sampler could effectively select frames containing key actions for each video.

	\subsubsection{Analysis of action alignment}
	\label{sec:abalation_align}
	\noindent
	\textbf{Design of the temporal alignment.} For the action alignment, we first analyze the effectiveness of our temporal coordination module by performing the temporal transformation and re-arrangement modules separately. In the summary of results in \autoref{tab:TC_design}, we can see that applying the single Temporal Transformation provides stable improvements across datasets ($\uparrow$ 1.7 accuracy on SSv2 and $\uparrow$ 3.8 on UCF101). The Temporal Rearrangement module brings more significant improvements to the baseline performance compared to the Temporal Transformation module. By applying both modules sequentially, we can reach the best performance, demonstrating that these two operations are essential to cope with different misalignment issues (duration and evolution misalignment) in the temporal dimension.

\begin{table}[t]
	\centering
	\begin{tabular}{l|cc}
		\hline
		\multicolumn{1}{c|}{Setting} & UCF101 & SSv2 \\ \hline
		Baseline (ProtoNet)                     &  78.7    &   34.2   \\ \hline
		+ Temporal Transformation    &  82.5    &   35.9   \\ \hline
		+ Temporal Re-arrangement    &  83.8   &   37.3   \\ \hline
		+ Both                       &  \textbf{84.2}    &    \textbf{37.9}  \\ \hline
	\end{tabular}
	\caption{Effectiveness of each module in TC, reported under 5-way 1-shot setting.}
	\label{tab:TC_design}
\end{table}

\begin{table}[t]
	\small
	\centering
	\begin{tabular}{c|cc}
		\hline
		Solution	& UCF101 & HMDB \\ \hline
		Enumerate & 84.9 & 59.6 \\
		Affine Transformation & 85.6 & 60.4 \\
		Mask-based (Our implementation) & \textbf{86.2} & \textbf{61.1} \\ \hline
	\end{tabular}
	\caption{Results of different implementation of SC in our VIM, reported under 5-way 1-shot setting.}
	\label{tab:SC_design}
	\vspace{-0.75em}
\end{table}	
	
	\noindent
	\textbf{Different designs for spatial alignment}
	Furthermore, to reveal the effectiveness of spatial coordination (SC) module design, we compare our implementation with other alternative designs:
	\begin{itemize}
		\item [(1)] \textit{Enumerate}: 
        it enumerates through all possible \textit{integer} offsets in $\mathbb{Z}^{2}$, and straightly indexes the intersection area in feature maps by $x,y$ coordinates. The distance between support and query features is calculated within this intersection area. The offset leading to the minimum distance is considered optimal. This exhaustive design can be regarded as a simple baseline.

		\item  [(2)] \textit{Affine Transformation}: 
        it generates a 2D grid (i.e., parameters that describe an affine transformation) according to our predicted offset. Then it performs corresponding affine transformation over the entire 2D feature map to obtain features in the intersection area with re-sampling. (similar to STN~\cite{STN2015}).
	\end{itemize}
	We report their results on two appearance-dominated datasets: UCF101 and HMDB in~\autoref{tab:SC_design}. As we can see, our implementation outperforms the simple enumeration method by a great margin, which also demonstrates the ability of our offset predictor in SC. Compared to the affine transformation implementation, our alignment design is definitely more straightforward and computationally tractable, with slightly better performance.
 
\begin{figure}[t]
	\centering
	\includegraphics[width=\linewidth]{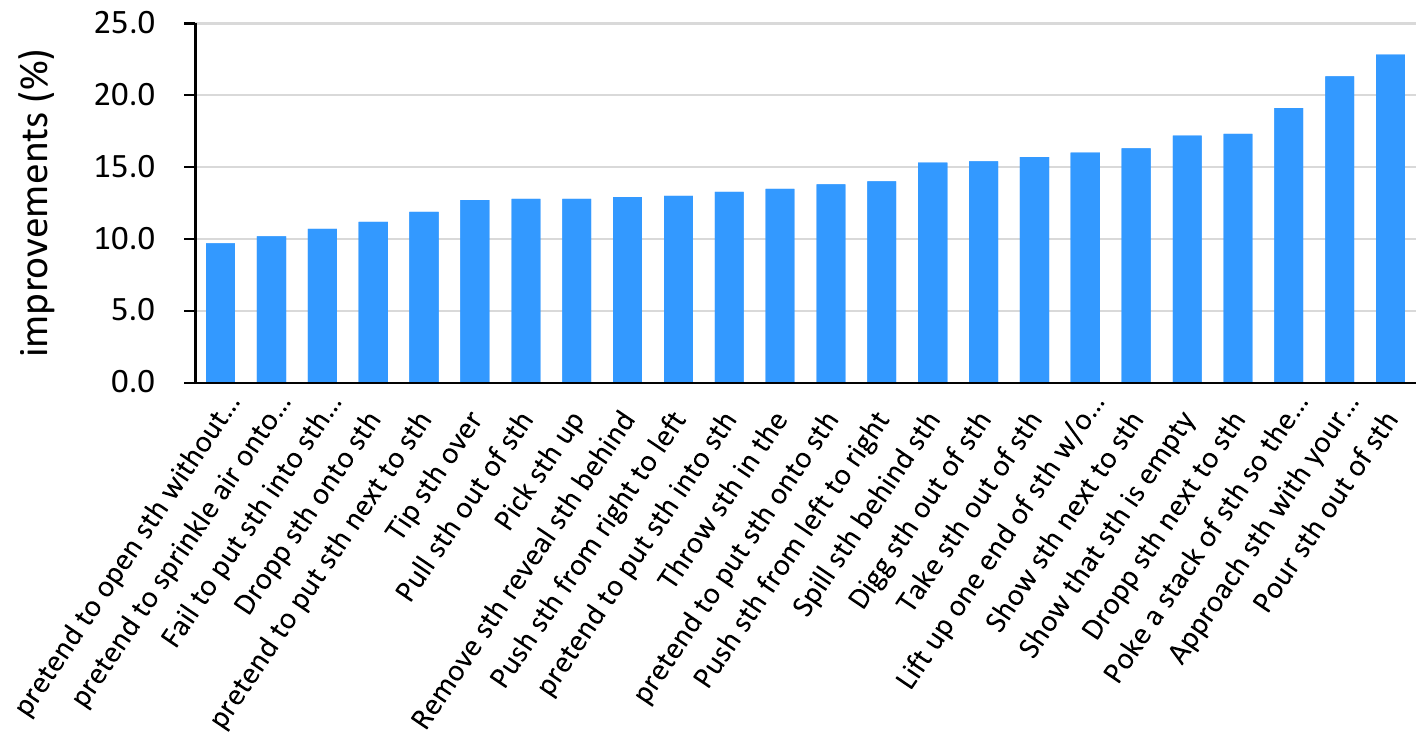}
	\caption{Class-specific improvement using our alignment model compared to the ProtoNet baseline in SSv2. Here we remove the adaptive sampling in VIM to analyze the gain from the action alignment independently.}
	\label{fig:class_improvement}
\end{figure}

	\noindent
	\textbf{Per-category improvement.} The improvement on the SSv2 dataset for each category using the proposed alignment model is presented in~\autoref{fig:class_improvement}. Note that here we remove the adaptive sampling in VIM to analyze the gain from the alignment independently. What stands out in this figure is that the performance increases by a large margin for all categories. Moreover, some categories' accuracy (\emph{e.g.} ``\textit{pouring sth out of sth}'', ``\textit{approaching sth}'', ``\textit{poking a stack of sth}'') rose sharply ($>20\%$ improvement) using our alignment. Interestingly, these action classes are also the ones that are most vulnerable to the misalignment problem. This further proves the effectiveness of our temporal alignment.

    \begin{table}[t]
    \centering
    \begin{tabular}{c|c|c}
    \hline
        & Temporal first (\textbf{default}) & Spatial First \\ \hline
    UCF      & 86.2           & 82.9 (\down{$\downarrow$3.3})         \\ \hline
    HMDB     & 61.1           & 58.7 (\down{$\downarrow$2.4})         \\ \hline
    Kinetics & 73.8           & 69.6 (\down{$\downarrow$4.2})         \\ \hline
    SSv2     & 40.8           & 38.5 (\down{$\downarrow$2.3})         \\ \hline
    \end{tabular}
    \caption{The results (1-shot) under different orders of alignment in VIM.}
    \label{tab:order_align}
    \end{table}
	
	\noindent
	\textbf{Temporal or spatial alignment first?}
	We further investigate the effect of alignment order by exchanging the order of the spatial coordination and temporal coordination modules in VIM.  The results are listed in \autoref{tab:order_align}, which suggests that the performance is significantly affected by the order.  Notably, the VIM achieves the highest performance across all datasets when temporal alignment is performed first, followed by spatial alignment.  Meanwhile, we also observe a sharp decline in performance on UCF and Kinetics when spatial alignment is conducted first. This decline can be attributed to the spatial coordination module's reliance on temporally consistent frame features to predict accurate spatial offsets. Based on these findings, we can conclude that a preliminary temporal alignment is essential for successful spatial alignment in VIM.

    \noindent
    \textbf{Comparison to sequence alignment methods.}
    To validate the effectiveness and superiority of our proposed alignment approach, we replace the action alignment procedure in VIM with other popular sequence alignment algorithms:
    \begin{itemize}
        \item Soft-DTW~\cite{softdtw}. It enables the Dynamic Time Warping (DTW) algorithm to be differentiable for end-to-end training in neural networks.
        \item Drop-DTW~\cite{drop-dtw}. To cope with the presence of outliers that can be arbitrarily interspersed in the sequences, it performs alignment while automatically dropping the outlier elements from the matching process.
        \item VAVA~\cite{align_wild}. It learns to align sequential actions in the wild that involve diverse temporal variations (e.g., background frames) by enforcing temporal priors on the optimal transport matrix.
    \end{itemize}
    We present the results of replacing our alignment model with the above methods in~\autoref{tab:compare_align}. Both of them can also relieve the misalignment issue and improve performance. VAVA surpasses the Soft-DTW and Drop-DTW in most cases thanks to its ability to handle noisy frames (e.g., background frames) during alignment. It's worth noting that these algorithms only align actions in the temporal dimension. In contrast, our framework performs action alignment in both temporal and spatial dimensions, resulting in the best overall performance compared to the other methods. Additionally, utilizing $\mathcal{L}_{\text{inter}}$ can further improve our performance, emphasizing the importance of explicit guidance for alignment.

    \begin{table}[t]
    \centering
    \resizebox{\linewidth}{!}{
    \begin{tabular}{c|cccc}
    \hline
    Action alignment in VIM & HMDB & UCF & Kinetics & SSv2 \\ \hline
    Soft-DTW~\cite{softdtw}                &  58.0   &  81.6    & 73.0 &  38.2   \\ \hline
    Drop-DTW~\cite{drop-dtw}                &  57.2   &  82.5    &  72.6 &  38.4 \\ \hline
    VAVA~\cite{align_wild}                  &  58.9   & 83.5    & 73.2 &  37.8  \\ \hline
    Ours (w/o $\mathcal{L}_{\text{inter}}$)                   &   60.6  &   84.6 & 73.6 & 39.0  \\ \hline
    Ours (with $\mathcal{L}_{\text{inter}}$)                   &   \textbf{61.1}  &   \textbf{ 86.2 } & \textbf{73.8 } & \textbf{40.8}  \\ \hline
    \end{tabular}}
    \caption{Results of replacing our proposed action alignment in VIM with other alignment algorithms. }
    \label{tab:compare_align}
    \end{table}
    
		\begin{figure}[t]
		\centering
\includegraphics[width=\linewidth]{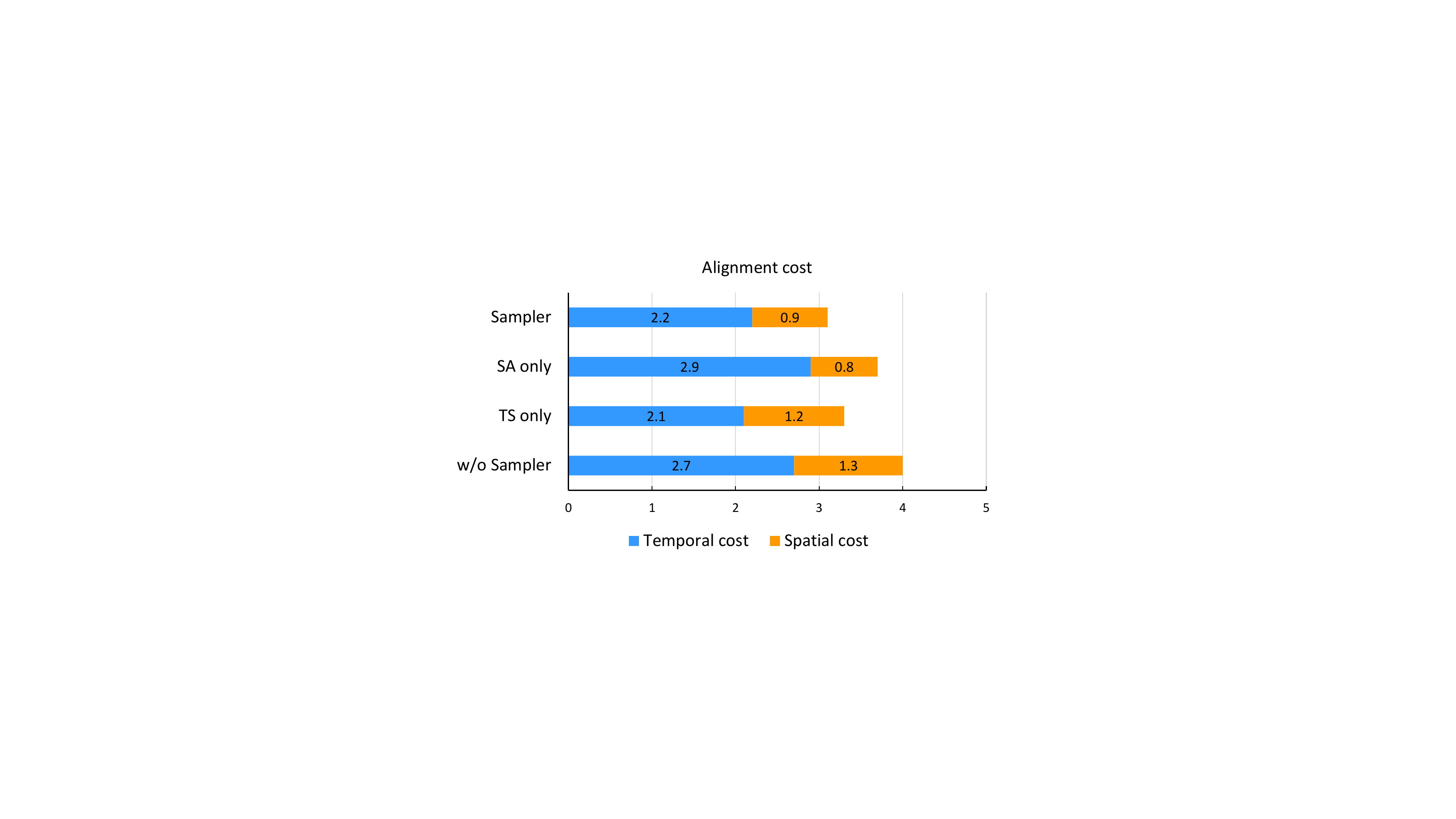}
		\caption{The alignment cost with different operations on input videos. `SA only' indicates that we only adopt spatial amplifier only. `TS' means we only perform temporal selector.}
		\label{fig:align_cost}
	\end{figure}

		\begin{figure}[t]
		\centering
\includegraphics[width=0.9\linewidth]{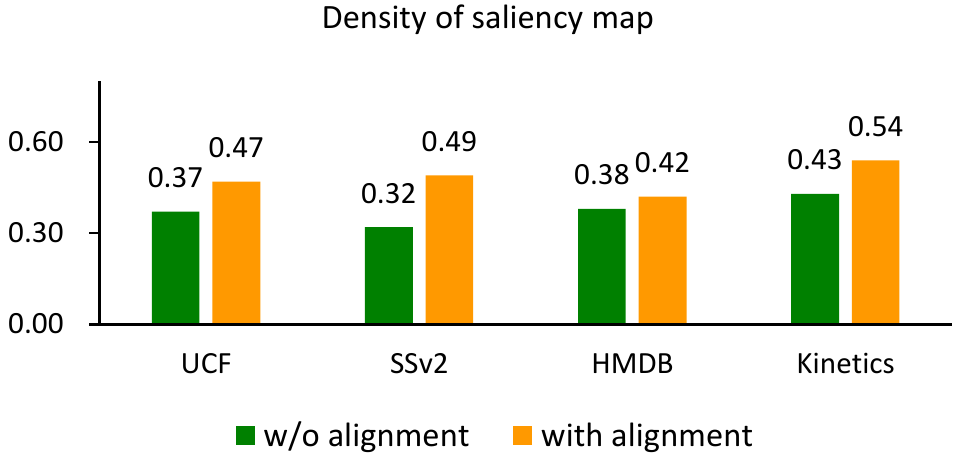}
		\caption{Density of saliency maps in spatial amplifier. A higher density indicates reduced amplification.}
		\label{fig:density}
	\end{figure}
 
	\subsubsection{Interaction between the sampler and alignment}
	\noindent
	\textbf{Impact on alignment cost.} We first delve into the effect of adaptive sampling on action alignment. To quantify the alignment process, we calculate the \textit{alignment cost} for each input video as follows:
	\begin{gather}
		\mathcal{C}_{\text{align}} = \underbrace{(|1-a|+|b|) + \lVert \mathbf{M}_{e}-I \rVert_2}_{\text{temporal cost}} + \underbrace{\lVert O \rVert}_{\text{spatial cost}} 
	\end{gather}
	where $(a,b)$ are the affine parameters for linear temporal transformation, $T$ is the number of frames, $\mathbf{M}_e \in \mathbb{R}^{T\times T}$ denotes the normalized evolution correlation matrix calculated in temporal re-arrangement (\autoref{eq:tam}), $I\in \mathbb{R}^{T\times T}$ is an identity matrix (indicates two videos are frame-wise consistent), $O\in \mathbb{R}^{T\times 2}$ indicates the predicted spatial offset in \autoref{eq:offset} (here we normalize values in $O$ into [-1,1]). In this context, a higher cost indicates a larger degree of misalignment in the input video sequences. It is worth noting that the alignment cost is determined based on the estimated parameters of our alignment model, rather than an exact measurement. Our results are presented with the mean value of $\mathcal{C}_{\text{align}}$, both with and without adaptive video sampling, and are summarized in Figure \ref{fig:align_cost}. The findings demonstrate that incorporating the sampler significantly reduces the overall alignment cost of our action alignment model. Furthermore, it is observed that the temporal alignment cost slightly increases when we only employ the spatial amplifier. 

    \noindent
    \textbf{Impact on sampling strategy.} We also investigate how the feature-level alignment operation impacts the behavior of the video sampler.   To quantify the behavior of our sampler, we compute the density (values between 0$\sim$1) of generated 2D saliency map $\mathbf{M}_s$ (\autoref{eq:saliency}).    In our spatial amplifier, a higher density corresponds to reduced amplification.  For example, a density of 1 implies equal importance for all pixels, and therefore no local region will be amplified.  As seen in \autoref{fig:density}, there is a slight increase in density when the alignment operation is incorporated at the feature level.  This suggests that the presence of spatial alignment allows for a less optimal solution for amplification, as the critical regions will be further aligned at the feature level.   On the other hand, the sampler is required to provide more precise and concentrated amplification (i.e., effectively highlighting the actor region) as input to subsequent networks when no alignment of spatial features is present.   Consequently, the existence of the alignment can alleviate the training difficulty faced by the video sampler.

	\noindent
	\textbf{Balance between intra- and inter-video information.} As presented in \autoref{eq:loss}, the coefficients of $\mathcal{L}_{\text{intra}}$ and $\mathcal{L}_{\text{inter}}$ control the preference of VIM for learning intra- and inter-video information during training, i.e., balance the optimization of the sampler and alignment. To evaluate the impact of these weights, we perform experiments using different ratios of $\mu_1, \mu_2$. Specifically, we set the values for $\mu_1,\mu_2$ to $\{1, 0\}, \{1, 0.5\}, \{0.5, 1\}, \{1, 1\}$, and $\{0, 1\}$, and train our VIM on each dataset using these pairs of coefficients. The results are presented in \autoref{tab:loss_weight}. Disabling either loss term (i.e., setting its coefficient to 0) leads to a rapid drop in performance.  When treating them equally during training, we can already reach an ideal performance. However, scaling the ratio of $\mathcal{L}_\text{inter}$  and $\mathcal{L}_\text{intra}$ to 2 (i.e., $\mu_1=0.5,\mu_2=1$) further boosts the performance on both UCF101 and SSv2 datasets, which demonstrates that the inter-video information may be more valuable for the final recognition in our VIM.

    \begin{table}[t]
    \centering
    \begin{tabular}{c|cc}
    \hline
    Coefficient of $\mathcal{L}_{\text{intra}}$ and $\mathcal{L}_{\text{inter}}$ & UCF & SSv2 \\ \hline
    $\mu_1=1,~\mu_2=0$   &  84.6   &  39.0    \\ \hline
    $\mu_1=0,~\mu_2=1$   &  83.5   &    40.3  \\ \hline
    $\mu_1=1,~\mu_2=0.5$   &   85.1  &   40.1   \\ \hline
    $\mu_1=0.5,~\mu_2=1$   &  \textbf{86.2}   &  \textbf{40.8}   \\ \hline
    $\mu_1=1,~\mu_2=1$   &   85.8  &   40.5   \\ \hline
    \end{tabular}
    \caption{The results of VIM under different coefficients of $\mathcal{L}_{\text{intra}}$ and $\mathcal{L}_{\text{inter}}$.}
    \label{tab:loss_weight}
    \end{table}

    \begin{table}[t]
    \centering
    \resizebox{\linewidth}{!}{%
    \begin{tabular}{l|c|c|c|c}
    \hline
    \multicolumn{1}{c|}{Setting}    & UCF  & HMDB & Kinetics & SSv2 \\ \hline
    VIM                     & 86.2 & 61.6 & 73.8     & 40.8 \\ \hline
    Remove sampler during testing   & 84.7 & 58.8 & 71.9     & 37.7 \\ \hline
    Remove alignment during testing & 82.6 & 58.1 & 71.1     & 34.3 \\ \hline
    Individual sampler              & 83.4 & 59.6 & 72.1     & 36.8 \\ \hline
    Individual alignment            & 85.4 & 60.4 & 73.0     & 39.6 \\ \hline
    \end{tabular}%
    }
    \caption{Results of test-time isolation of VIM.}
    \label{tab:isolation}
    \end{table}
 
	\noindent
	\textbf{Test-time isolation.}
	We further conducted experiments to analyze the interdependence between the sampler and alignment modules in our VIM.  We removed either the sampler or alignment module during test time while keeping both during training, a process we call ``test-time isolation''.  The results are presented in \autoref{tab:isolation}, where we observe a significant drop in the performance of VIM when either module was removed during testing. Most importantly, the performance of the test-time isolated models is also lower than that of individually trained sampler or alignment models (refer to \autoref{tab:breakdown1}). These findings indicate that optimizing the sampler and alignment jointly in our VIM does encourage them to work together and be deeply coupled, rather than a simple incremental stack.

	\begin{table}[t]
		\centering
		\begin{tabular}{c|cc|cc}
			\hline
			\multirow{2}{*}{Models}   & \multicolumn{2}{c|}{Setting} & \multirow{2}{*}{UCF101} & \multirow{2}{*}{SSv2} \\ \cline{2-3}
			& Number          & Resolution             &                         &                       \\ \hline
			\multirow{4}{*}{ProtoNet} & 8          & 224$\times$224         & 78.7                    & 34.2                  \\ \cline{2-5} 
			& 8          & 448$\times$448         & 79.9                    & 34.7                  \\ \cline{2-5} 
			& 20         & 224$\times$224         & 80.8                    & 35.6                  \\ \cline{2-5} 
			& 20         & 448$\times$448         & 82.1                    & 36.3                  \\ \hline
			+ Our sampler             & 8          & 224$\times$224         & \textbf{83.4}                    & \textbf{37.8}                  \\ \hline
		\end{tabular}
		\caption{Comparison with the estimated upper bound of intra-video information. Results are reported under 1-shot setting.}
		\label{tab:upper}
	\end{table}

    \begin{table}[t]
    \centering
    \resizebox{\linewidth}{!}{
    \begin{tabular}{c|l|cc}
    \hline
    Model                     & \multicolumn{1}{c|}{Setting}  & UCF  & SSv2 \\ \hline
    \multirow{7}{*}{ProtoNet} & \multicolumn{1}{l|}{Baseline} & 78.7 & 34.2 \\ \cline{2-4} 
                              & + Temporal Exhaustive          & 82.3 & 36.5 \\
                              & + Spatial Exhaustive           & 82.7 & 35.3 \\
                              & + Both                         & 83.9 & 37.1 \\ \cline{2-4} 
                              & + Our Temporal Coordination        & 84.2 & 37.9 \\
                              & + Our Spatial Coordination         & 83.3 & 36.0 \\
                              & + Both (Our complete alignment)                         & \textbf{85.4} & \textbf{39.6} \\ \hline
    \end{tabular}}
    \caption{Comparison with the estimated upper bound of inter-video information. Results are reported under 1-shot setting.}
    \label{tab:upper_inter}
    \end{table}

\begin{figure*}[t]
\centering
\begin{minipage}[t]{0.45\textwidth}
\centering
\includegraphics[width=\textwidth]{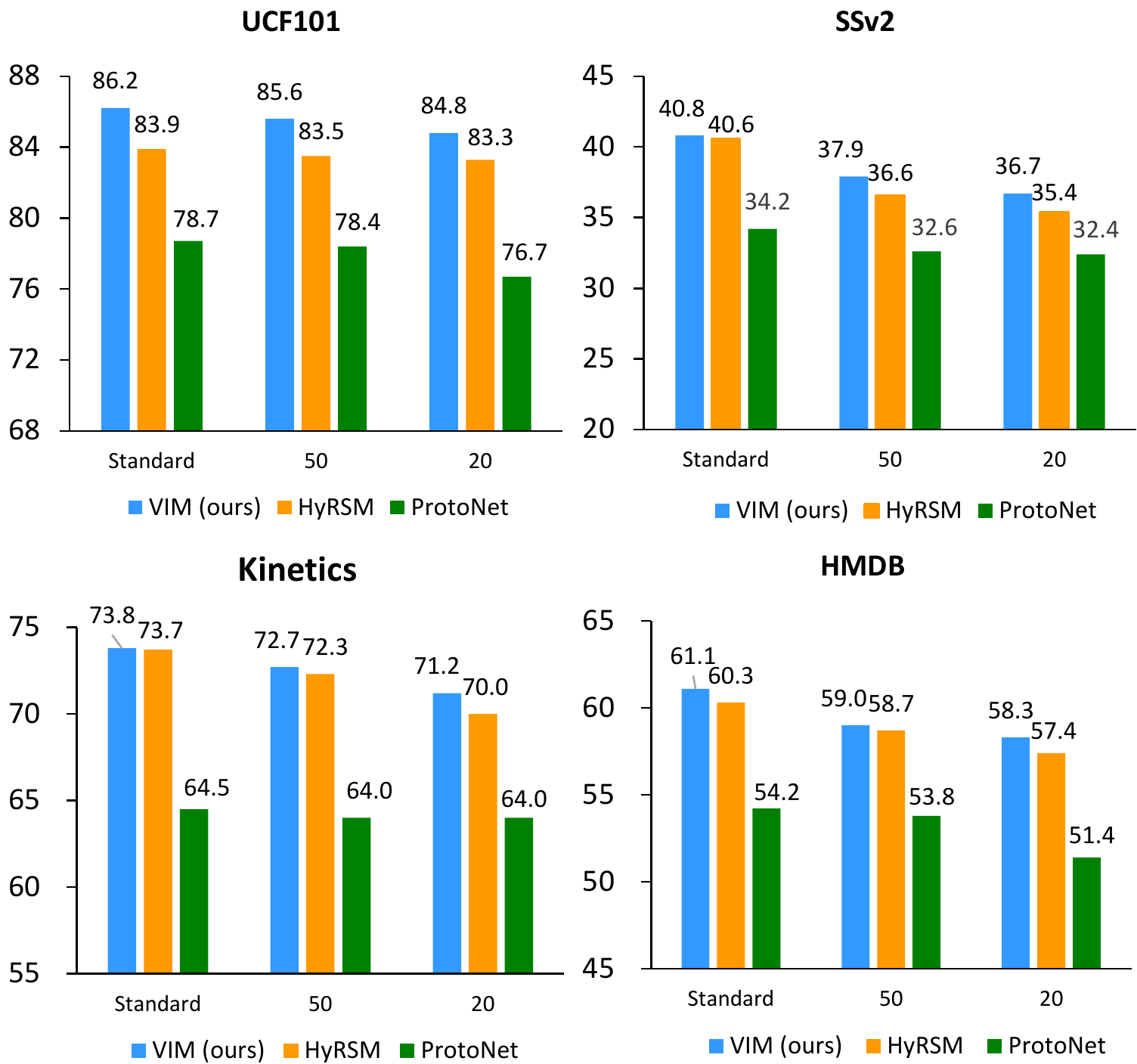}
\caption{Results of the limit test. ``\textit{standard}'' indicates the standard few-shot data split.}
\label{fig:limit_test}
\end{minipage}
\hspace{2pt}
\begin{minipage}[t]{0.48\textwidth}
\centering
\includegraphics[width=\textwidth]{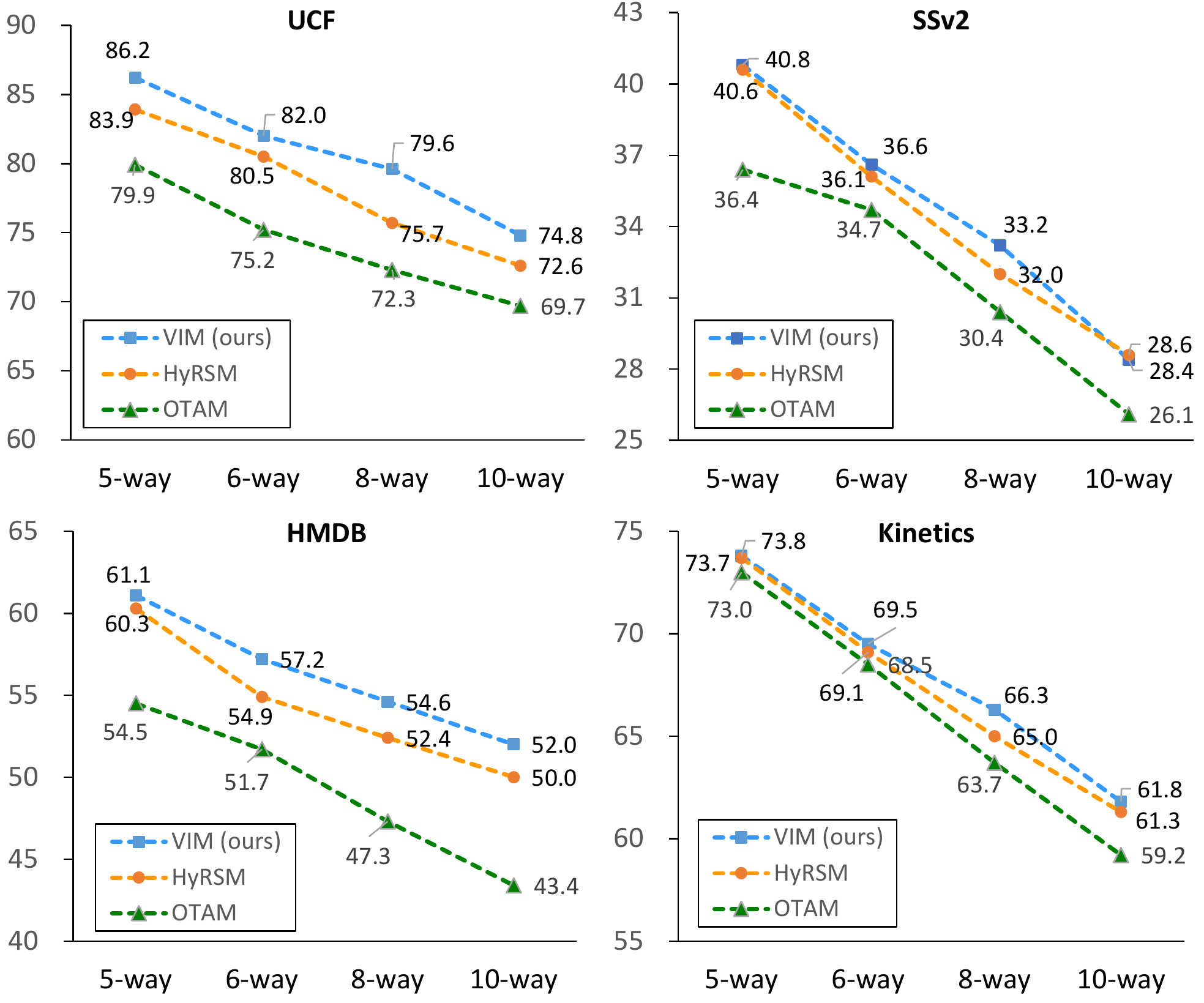}
\caption{Results of many-way 1-shot setting. The number of ways ranges from 5 to 10.}
\label{fig:many_way}
\end{minipage}
\end{figure*}

	\subsubsection{Probe the upper bound of video information}
	
	\noindent
	\textbf{Intra-video information.} 
    As previously discussed, increasing the number and resolution of input frames can directly improve the coverage and utilization of video data for few-shot learners.  To further investigate the effectiveness of our sampler, we follow this naive solution to estimate a rough upper bound of intra-video information at the data level.  Results are summarized in ~\autoref{tab:upper}.  Although this approach incurs significant GPU memory and computational costs, we can observe consistent improvements by increasing the frame number and resolution.  Specifically, the performance under 5-way 1-shot with $T=20$ and $H=W=448$ (the max values we can afford for training) can be considered an upper bound of intra-video information at the data level.  However, when integrating the baseline with our proposed sampler, its performance can \textit{\textbf{surpass}} the upper bound by a significant margin without enlarging the frame number or resolution.  This also demonstrates that this naive solution \textit{does work but is not ideal} compared to our adaptive sampler.  
	
	\noindent
	\textbf{Inter-video information.} The rough upper bound of inter-video information can be estimated in an exhaustive alignment manner. As for the temporal dimension, we imitate TRX~\cite{trx2021} to construct exhaustive representations of pairs/triplets of frames following the temporal order. The metric learning between two video features is conducted over all possible representations, and we calculate a mean value as the final distance. As for the spatial dimension, we adopt the enumerate design (as described in the 2nd paragraph of \autoref{sec:abalation_align}) to estimate the upper bound results.  Moreover, we add more \textit{non-integer} offsets by interpolating between the possible integer offsets to cover more candidates.  In the same way, after calculating metric distances with all enumerated offsets, we regard the minimum distance as the final result. The results of the above exhaustive alignment and our alignment model are summarized in \autoref{tab:upper_inter}. Exhaustive alignment provides stable improvements across datasets, especially on UCF101, as seen in our results.  However, our proposed alignment still outperforms it in all cases, demonstrating that our alignment model can find good alignments using a limited amount of training data.  Although the improvements provided by spatial exhaustive alignment are comparable to our spatial consistency (SC) method, our SC is more flexible and learnable.

	\subsubsection{Limit test with extremely fewer data}
	To further validate the effectiveness of our VIM, we conducted a limit test using an extremely limited amount of video data for training. In standard few-shot data splits, each action category typically contains approximately 100-130 samples for training. In the limit test, we reduced this number further to 50 and 20 to analyze the performance of our VIM and other methods under extremely challenging circumstances. The results are summarized in~\autoref{fig:limit_test}. Our VIM still delivered significant improvements over the baseline in these extreme circumstances. Although the HyRSM~\cite{HyRSM2022} performs comparably to our VIM in Kinetics and SSv2 datasets under standard settings, our VIM outperforms it by a remarkable margin when training with limited data. These findings indicate that our VIM is more robust and effective than existing methods when faced with extremely limited training data.

	\subsubsection{Many-way few-shot}
	We have extended the conventional 5-way few-shot setting, which consists of 5 categories in each episode, to a more adaptable \textit{many-way} few-shot setting. In \autoref{fig:many_way}, we present the 1-shot results from 6-way to 10-way. It is evident that as the number of ways increases, it becomes more challenging for the classifier to distinguish among more action categories, resulting in a decline in performance. Nevertheless, our proposed VIM consistently surpasses current state-of-the-art methods such as HyRSM~\cite{HyRSM2022} and OTAM~\cite{otam2020}. These results highlight the superiority of our VIM in handling diverse scenarios, enabling it to generalize effectively.

  \begin{figure*}
 \centering
    \begin{minipage}[t]{0.48\textwidth}
        \centering
        \resizebox{\linewidth}{!}{
        \begin{tabular}{c|c|c|c|c|c}
        \hline
        \begin{tabular}[c]{@{}c@{}}Training \\ domain\end{tabular} & \begin{tabular}[c]{@{}c@{}}Test \\ domain\end{tabular} & ProtoNet & OTAM~\cite{otam2020} & HyRSM~\cite{HyRSM2022} & VIM \\ \hline
        UCF      & UCF  & 78.7 & 79.9 & 83.9 & \textbf{86.2} \\
        HMDB     & HMDB & 54.2 & 54.5 & 60.3 & \textbf{61.1} \\
        SSv2     & SSv2 & 34.2 & 36.4 & 40.6 &  \textbf{40.8}\\ \hline
        Kinetics & UCF  & 78.5 & 79.7 & 83.1 & \textbf{83.5} \\
        Kinetics & HMDB & 53.9 & 54.0 & 56.7 & \textbf{58.1} \\
        Kinetics & SSv2 & 31.6 & 30.6 & 32.0 & \textbf{32.4} \\ \hline
        \end{tabular}
        }
        \caption{The ability of domain generalization. Results are reported under 1-shot setting.}
        \label{tab:domain}
    \end{minipage}
    \hspace{10pt}
    \begin{minipage}[t]{0.48\textwidth}
		\centering
		\small
		\begin{tabular}{c|c|c|c}
			\hline
			Model            & Params & Inference time  & Performance     \\ \hline
			ProtoNet         & 25.8 M & 0.67 s/task & 78.7 \\ \hline
            OTAM~\cite{otam2020} & 25.8 M  & 1.29 s/task & 79.9 \\
            TRX~\cite{trx2021} & 47.1 M  & 0.92 s/task  & 78.6 \\
            HyRSM~\cite{HyRSM2022} & 65.6 M  & 0.89 s/task & 83.9 \\
            \hline
			VIM (Ours) & 58.2 M  & 0.85 s/task & \textbf{86.2} \\ 
            \hline
		\end{tabular}
		\caption{Complexity analysis. The results are reported under 5-way 1-shot setting on UCF101 dataset. Inference time is tested on a single NVIDIA V100 GPU.}
		\label{tab:overhead}
    \end{minipage}
 \end{figure*}

  	\begin{figure*}[t]
		\centering
		\includegraphics[width=1.0\linewidth]{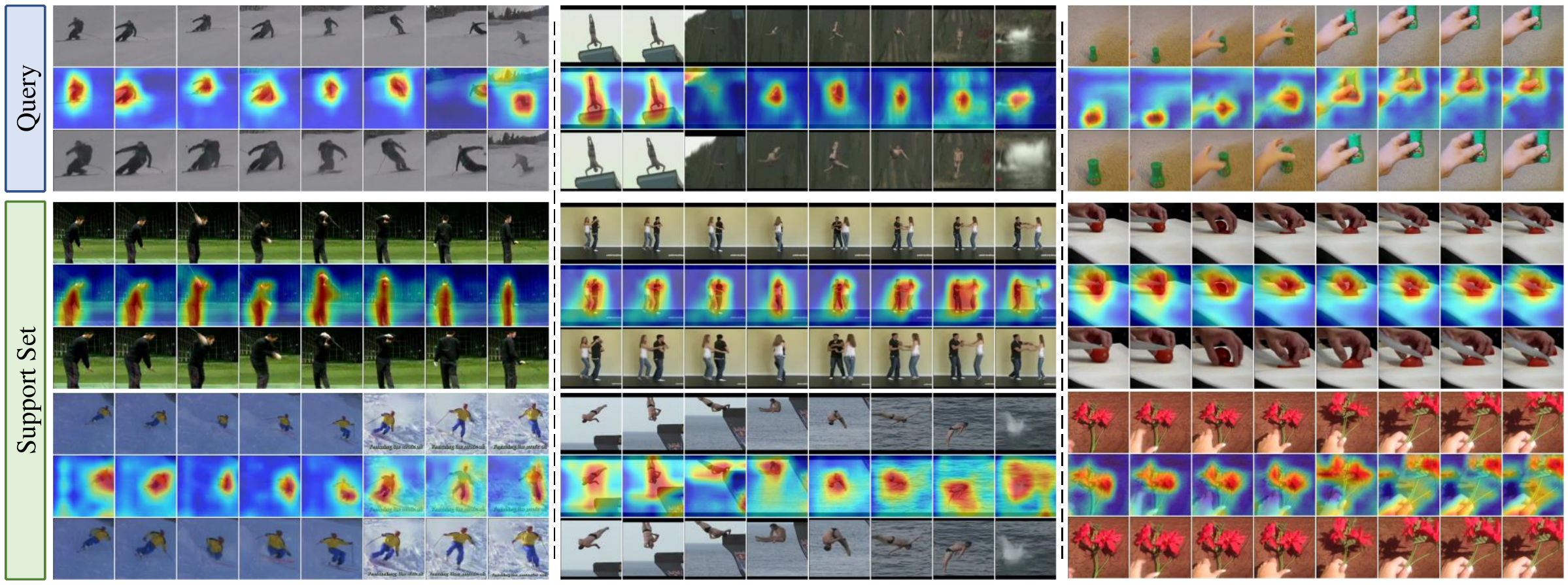}
		\caption{Visualization results of sampler in our VIM in three sample episode tasks from UCF and SSv2 datasets. The \textit{1st} row: frames selected by Temporal Selector (TS), the \textit{2nd} row: saliency maps, the 3rd row: frames amplified by Spatial Amplifier (SA). 2-way 1-shot is illustrated here.
		}
		\label{fig:vis_sampler}
	\end{figure*}

 \begin{figure*}[t]
		\centering
		\includegraphics[width=1.0\linewidth]{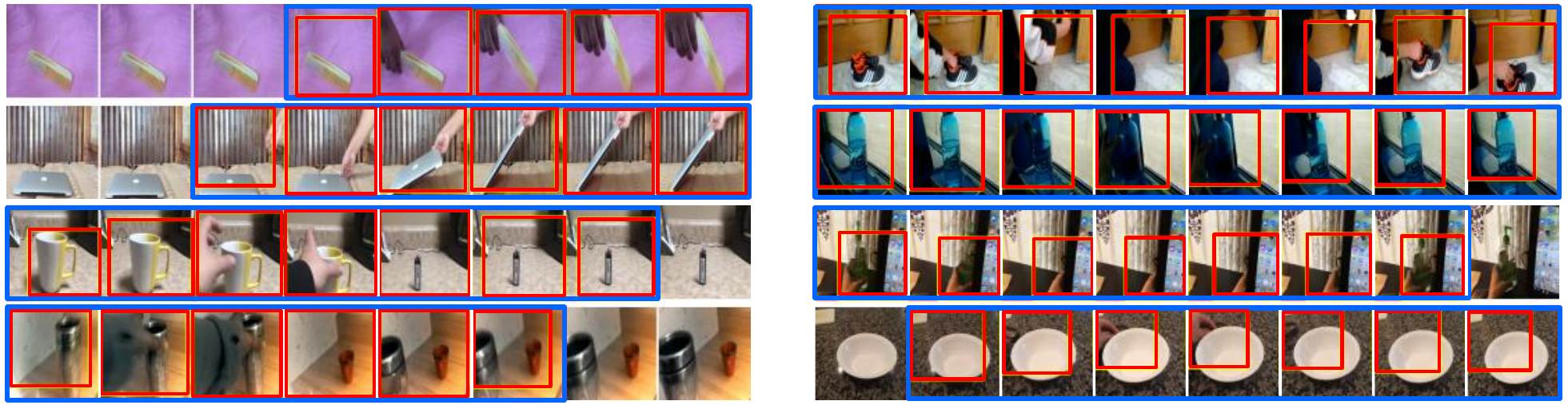}
		\caption{Visualization results of action alignment on SSv2. Action duration (highlight with blue boxes) is obtained by predicted warp parameters in Temporal Coordination (TC). Action-specific spatial region (highlight with red box) is located with predicted offset in Spatial Coordination (SC). 
		}
		\label{fig:vis_align}
	\end{figure*}
 \begin{figure*}[ht]
	\centering
	\begin{minipage}[t]{1.0\textwidth}
		\centering
		\includegraphics[width=1\textwidth]{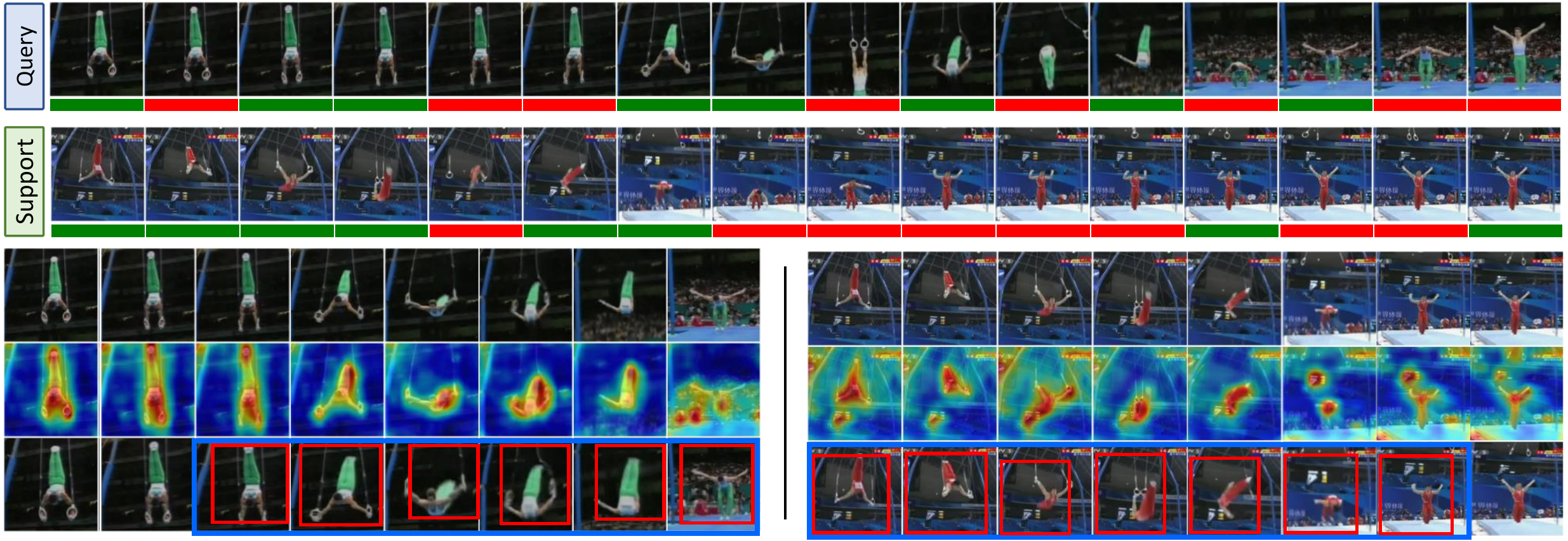}
		\subcaption{}
	\end{minipage}
	\begin{minipage}[t]{1.0\textwidth}
		\centering
		\includegraphics[width=1\textwidth]{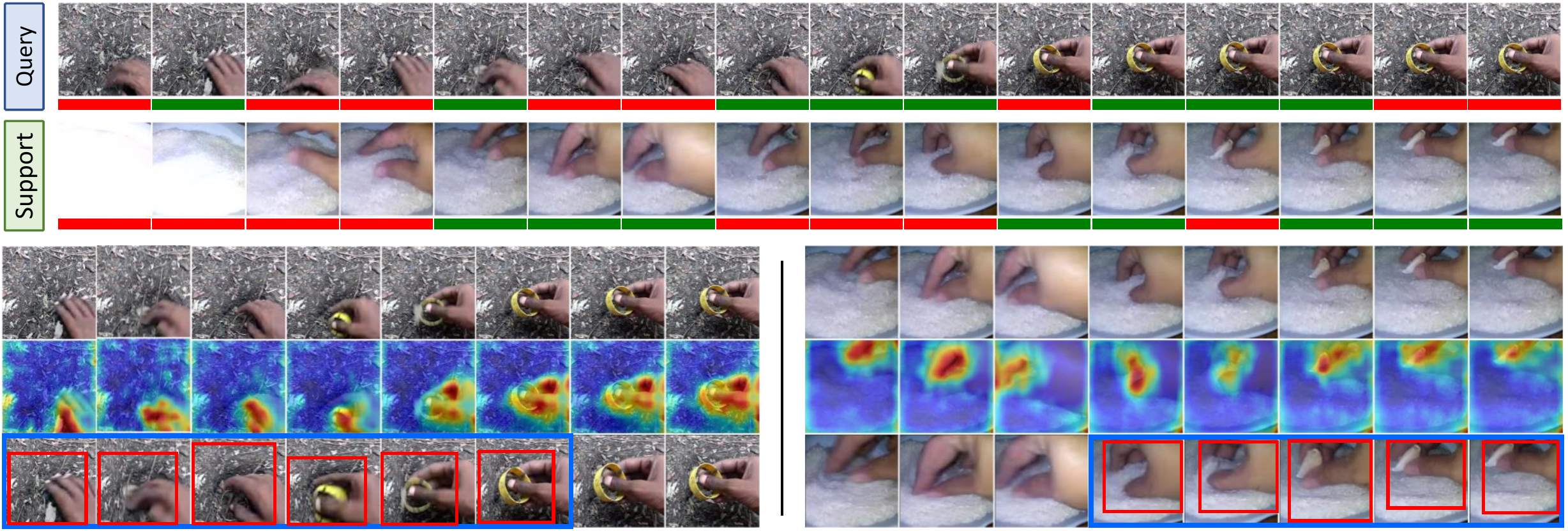}
		\subcaption{}
	\end{minipage}
	\caption{Visualization results of the complete VIM on selected videos on UCF101 and SSv2 datasets. Selected frames are indicated by green bar, while other results are presented in the same way as  \autoref{fig:vis_sampler} and    \autoref{fig:vis_align}.}
	\label{fig:vis_vim}
\end{figure*}
	\subsubsection{Complexity analysis} We further analyze the complexity and overhead of our VIM.
	As shown in \autoref{tab:overhead}, although our VIM introduces additional parameters (32.4M) to the baseline (ProtoNet), our inference speed is still fast compared to current state-of-the-art methods. In particular, OTAM~\cite{otam2020} incurs heavy time costs due to its dynamic programming operation.  TRX~\cite{trx2021} requires a temporally exhaustive feature comparison, leading to a relatively slow inference speed. HyRSM~\cite{HyRSM2022} has the largest number of parameters as it includes many high-dimensional linear transformation layers. In comparison, our VIM model is faster and achieves the best performance.

	\subsubsection{Ability of domain generalization}
	To evaluate the domain generalization ability of VIM, we investigated whether few-shot learners trained on a single dataset could be generalized to other domains without additional fine-tuning.    The results are summarized in~\autoref{tab:domain}. They show that our Kinetics pre-trained VIM can generalize well to other domains and achieve ideal performance, particularly on UCF and HMDB datasets.  Moreover, we observed that this domain transfer paradigm could achieve comparable performance to standard training for various methods in the UCF and HMDB datasets.  This can be attributed to the similarity in domain knowledge between the Kinetics dataset and UCF and HMDB. In contrast, due to the significant domain gap between Kinetics and SSv2, the improvement from Kinetics to SSv2 is limited, and its performance is significantly behind that of standard training. However, our VIM still provides the largest improvement under this circumstance compared to other methods.

	\subsubsection{Visualization results}
    \noindent
    \textbf{Video sampler.}  We first visualize the temporal-spatial sampling results for videos in \autoref{fig:vis_sampler}. We observe that TS can select frames containing crucial and complete action processes. Moreover, our generated salience map complements by indicating the important spatial regions for each frame. Accordingly, as the results of SA show, these regions with higher saliency are well amplified and emphasized, which in turn, makes them easier to be recognized. Meanwhile, the amplified images still maintain relatively complete information of the original ones. In summary, the visualization tellingly depicts the effectiveness of our sampling strategy in temporal and spatial dimensions.

    \noindent
	\textbf{Action alignment.} We further visualize the alignment results to illustrate the effectiveness of our proposed method, which are presented in \autoref{fig:vis_align}. It can be observed that there exists a clear duration and spatial-temporal evolution misalignment between the support and query videos. It’s clear that the duration is well-aligned by the TC, which filters the insignificant background frame noise. Besides, the spatial regions coordinated by SC focus on the common action-specific part between paired frames. For example, the region of hand (row 3 col 2) and the action-specific object `cup’ (row 3 col 1) can be located precisely, which leads to a well-aligned spatial evolution for videos before being compared.

    \noindent
    \textbf{Complete VIM.} Finally, we visualize the results of complete VIM in~\autoref{fig:vis_vim}. It can be seen that the video sampler could preserve and emphasize the informative parts of videos, then actions in paired videos are further aligned in both spatial and temporal dimensions. This observation proves that adaptive sampling and action alignment steps are complementary to each other, which works hand-in-hand to maximize the intra- and inter-video information in our VIM framework.

	\section{Conclusion}
    This paper presents a novel framework called Video Information Maximization (VIM) for few-shot action recognition. VIM aims to maximize both intra- and inter-video information by utilizing adaptive sampling and action alignment, respectively. Adaptive sampling is achieved through a video sampler that selects important frames and amplifies critical spatial regions based on the task at hand. It is able to preserve and emphasize informative spatiotemporal parts of video clips, while eliminating interference in the input data level. The action alignment performs temporal and spatial alignment sequentially at the feature level, resulting in a more precise measurement of inter-video similarity. To further optimize VIM, the framework incorporates auxiliary loss terms based on mutual information measurement during training, which provide concrete guidance for intra- and inter-video information learning. Experimental results demonstrate that VIM effectively improves few-shot recognition performance and achieves state-of-the-art results in most cases.


	
	

	%
	%

	\bibliographystyle{spmpsci}      
	\bibliography{egbib}   

\begin{thebibliography}{10}
\providecommand{\url}[1]{{#1}}
\providecommand{\urlprefix}{URL }
\expandafter\ifx\csname urlstyle\endcsname\relax
  \providecommand{\doi}[1]{DOI~\discretionary{}{}{}#1}\else
  \providecommand{\doi}{DOI~\discretionary{}{}{}\begingroup
  \urlstyle{rm}\Url}\fi

\bibitem{mi_representation}
Bachman, P., Hjelm, R.D., Buchwalter, W.: Learning representations by
  maximizing mutual information across views.
\newblock Advances in neural information processing systems \textbf{32} (2019)

\bibitem{mine}
Belghazi, M.I., Baratin, A., Rajeshwar, S., Ozair, S., Bengio, Y., Courville,
  A., Hjelm, D.: Mutual information neural estimation.
\newblock In: International conference on machine learning, pp. 531--540. PMLR
  (2018)

\bibitem{multual_subset2}
Beraha, M., Metelli, A.M., Papini, M., Tirinzoni, A., Restelli, M.: Feature
  selection via mutual information: New theoretical insights.
\newblock In: 2019 international joint conference on neural networks (IJCNN),
  pp. 1--9. IEEE (2019)

\bibitem{perturb2020}
Berthet, Q., Blondel, M., Teboul, O., Cuturi, M., Vert, J.P., Bach, F.:
  Learning with differentiable pertubed optimizers.
\newblock Advances in neural information processing systems \textbf{33},
  9508--9519 (2020)

\bibitem{tarn2019}
Bishay, M., Zoumpourlis, G., Patras, I.: Tarn: Temporal attentive relation
  network for few-shot and zero-shot action recognition.
\newblock arXiv preprint arXiv:1907.09021  (2019)

\bibitem{activity2015}
Caba~Heilbron, F., Escorcia, V., Ghanem, B., Carlos~Niebles, J.: Activitynet: A
  large-scale video benchmark for human activity understanding.
\newblock In: Proceedings of the ieee conference on computer vision and pattern
  recognition, pp. 961--970 (2015)

\bibitem{otam2020}
Cao, K., Ji, J., Cao, Z., Chang, C.Y., Niebles, J.C.: Few-shot video
  classification via temporal alignment.
\newblock In: Proceedings of the IEEE/CVF Conference on Computer Vision and
  Pattern Recognition, pp. 10618--10627 (2020)

\bibitem{I3d2017}
Carreira, J., Zisserman, A.: Quo vadis, action recognition? a new model and the
  kinetics dataset.
\newblock In: proceedings of the IEEE Conference on Computer Vision and Pattern
  Recognition, pp. 6299--6308 (2017)

\bibitem{differential2021}
Cordonnier, J.B., Mahendran, A., Dosovitskiy, A., Weissenborn, D., Uszkoreit,
  J., Unterthiner, T.: Differentiable patch selection for image recognition.
\newblock In: Proceedings of the IEEE/CVF Conference on Computer Vision and
  Pattern Recognition, pp. 2351--2360 (2021)

\bibitem{autoaug2018}
Cubuk, E.D., Zoph, B., Mane, D., Vasudevan, V., Le, Q.V.: Autoaugment: Learning
  augmentation policies from data.
\newblock arXiv preprint arXiv:1805.09501  (2018)

\bibitem{softdtw}
Cuturi, M., Blondel, M.: Soft-dtw: a differentiable loss function for
  time-series.
\newblock In: International conference on machine learning, pp. 894--903. PMLR
  (2017)

\bibitem{inverse_sample1986}
Devroye, L.: Sample-based non-uniform random variate generation.
\newblock In: Proceedings of the 18th conference on Winter simulation, pp.
  260--265 (1986)

\bibitem{drop-dtw}
Dvornik, M., Hadji, I., Derpanis, K.G., Garg, A., Jepson, A.: Drop-dtw:
  Aligning common signal between sequences while dropping outliers.
\newblock Advances in Neural Information Processing Systems \textbf{34},
  13782--13793 (2021)

\bibitem{static2016}
Edwards, H., Storkey, A.: Towards a neural statistician.
\newblock arXiv preprint arXiv:1606.02185  (2016)

\bibitem{finn2017model}
Finn, C., Abbeel, P., Levine, S.: Model-agnostic meta-learning for fast
  adaptation of deep networks.
\newblock In: International conference on machine learning, pp. 1126--1135.
  PMLR (2017)

\bibitem{racnn2017}
Fu, J., Zheng, H., Mei, T.: Look closer to see better: Recurrent attention
  convolutional neural network for fine-grained image recognition.
\newblock In: Proceedings of the IEEE conference on computer vision and pattern
  recognition, pp. 4438--4446 (2017)

\bibitem{AmeFu2020}
Fu, Y., Zhang, L., Wang, J., Fu, Y., Jiang, Y.G.: Depth guided adaptive
  meta-fusion network for few-shot video recognition.
\newblock In: Proceedings of the 28th ACM International Conference on
  Multimedia, pp. 1142--1151 (2020)

\bibitem{frameexit2021}
Ghodrati, A., Bejnordi, B.E., Habibian, A.: Frameexit: Conditional early
  exiting for efficient video recognition.
\newblock In: Proceedings of the IEEE/CVF Conference on Computer Vision and
  Pattern Recognition, pp. 15608--15618 (2021)

\bibitem{ssv2}
Goyal, R., Ebrahimi~Kahou, S., Michalski, V., Materzynska, J., Westphal, S.,
  Kim, H., Haenel, V., Fruend, I., Yianilos, P., Mueller-Freitag, M., et~al.:
  The" something something" video database for learning and evaluating visual
  common sense.
\newblock In: Proceedings of the IEEE International Conference on Computer
  Vision, pp. 5842--5850 (2017)

\bibitem{mutual_subset1}
Guo, B., Nixon, M.S.: Gait feature subset selection by mutual information.
\newblock IEEE Transactions on Systems, MAN, and Cybernetics-part a: Systems
  and Humans \textbf{39}(1), 36--46 (2008)

\bibitem{fsl-augment2}
Hariharan, B., Girshick, R.: Low-shot visual recognition by shrinking and
  hallucinating features.
\newblock In: Proceedings of the IEEE International Conference on Computer
  Vision, pp. 3018--3027 (2017)

\bibitem{he2016deep}
He, K., Zhang, X., Ren, S., Sun, J.: Deep residual learning for image
  recognition.
\newblock In: Proceedings of the IEEE conference on computer vision and pattern
  recognition, pp. 770--778 (2016)

\bibitem{deep_infomax}
Hjelm, R.D., Fedorov, A., Lavoie-Marchildon, S., Grewal, K., Bachman, P.,
  Trischler, A., Bengio, Y.: Learning deep representations by mutual
  information estimation and maximization.
\newblock arXiv preprint arXiv:1808.06670  (2018)

\bibitem{STN2015}
Jaderberg, M., Simonyan, K., Zisserman, A., Kavukcuoglu, K.: Spatial
  transformer networks.
\newblock arXiv preprint arXiv:1506.02025  (2015)

\bibitem{vae2013}
Kingma, D.P., Welling, M.: Auto-encoding variational bayes.
\newblock arXiv preprint arXiv:1312.6114  (2013)

\bibitem{survery_action}
Kong, Y., Fu, Y.: Human action recognition and prediction: A survey.
\newblock International Journal of Computer Vision \textbf{130}(5), 1366--1401
  (2022)

\bibitem{HMDB}
Kuehne, H., Jhuang, H., Garrote, E., Poggio, T., Serre, T.: Hmdb: a large video
  database for human motion recognition.
\newblock In: 2011 International conference on computer vision, pp. 2556--2563.
  IEEE (2011)

\bibitem{protogan2019}
Kumar~Dwivedi, S., Gupta, V., Mitra, R., Ahmed, S., Jain, A.: Protogan: Towards
  few shot learning for action recognition.
\newblock In: Proceedings of the IEEE/CVF International Conference on Computer
  Vision Workshops, pp. 0--0 (2019)

\bibitem{LGMNet2019}
Li, H., Dong, W., Mei, X., Ma, C., Huang, F., Hu, B.G.: Lgm-net: Learning to
  generate matching networks for few-shot learning.
\newblock In: International conference on machine learning, pp. 3825--3834.
  PMLR (2019)

\bibitem{ta2n2022}
Li, S., Liu, H., Qian, R., Li, Y., See, J., Fei, M., Yu, X., Lin, W.: Ta2n:
  Two-stage action alignment network for few-shot action recognition.
\newblock In: Proceedings of the AAAI Conference on Artificial Intelligence, 2,
  pp. 1404--1411 (2022)

\bibitem{keypoint_comp2020}
Lin, W., He, X., Dai, W., See, J., Shinde, T., Xiong, H., Duan, L.: Key-point
  sequence lossless compression for intelligent video analysis.
\newblock IEEE MultiMedia \textbf{27}(3), 12--22 (2020)

\bibitem{hieve2020}
Lin, W., Liu, H., Liu, S., Li, Y., Qian, R., Wang, T., Xu, N., Xiong, H., Qi,
  G.J., Sebe, N.: Human in events: A large-scale benchmark for human-centric
  video analysis in complex events.
\newblock arXiv preprint arXiv:2005.04490  (2020)

\bibitem{scapnet2021}
Liu, H., Li, J., Li, D., See, J., Lin, W.: Learning scale-consistent attention
  part network for fine-grained image recognition.
\newblock IEEE Transactions on Multimedia  (2021)

\bibitem{sampler2022}
Liu, H., Lv, W., See, J., Lin, W.: Task-adaptive spatial-temporal video sampler
  for few-shot action recognition.
\newblock In: Proceedings of the 30th ACM International Conference on
  Multimedia, pp. 6230--6240 (2022)

\bibitem{align_wild}
Liu, W., Tekin, B., Coskun, H., Vineet, V., Fua, P., Pollefeys, M.: Learning to
  align sequential actions in the wild.
\newblock In: Proceedings of the IEEE/CVF Conference on Computer Vision and
  Pattern Recognition, pp. 2181--2191 (2022)

\bibitem{shufflenet2018}
Ma, N., Zhang, X., Zheng, H.T., Sun, J.: Shufflenet v2: Practical guidelines
  for efficient cnn architecture design.
\newblock In: Proceedings of the European conference on computer vision (ECCV),
  pp. 116--131 (2018)

\bibitem{arnet2020}
Meng, Y., Lin, C.C., Panda, R., Sattigeri, P., Karlinsky, L., Oliva, A.,
  Saenko, K., Feris, R.: Ar-net: Adaptive frame resolution for efficient action
  recognition.
\newblock In: European Conference on Computer Vision, pp. 86--104. Springer
  (2020)

\bibitem{ATA2022}
Nguyen, K.D., Tran, Q.H., Nguyen, K., Hua, B.S., Nguyen, R.: Inductive and
  transductive few-shot video classification via appearance and temporal
  alignments.
\newblock In: Computer Vision--ECCV 2022: 17th European Conference, Tel Aviv,
  Israel, October 23--27, 2022, Proceedings, Part XX, pp. 471--487. Springer
  (2022)

\bibitem{saliencysampling2022}
Obeso, A.M., Benois-Pineau, J., V{\'a}zquez, M.S.G., Acosta, A.{\'A}.R.: Visual
  vs internal attention mechanisms in deep neural networks for image
  classification and object detection.
\newblock Pattern Recognition \textbf{123}, 108411 (2022)

\bibitem{metaUVFS2021}
Patravali, J., Mittal, G., Yu, Y., Li, F., Chen, M.: Unsupervised few-shot
  action recognition via action-appearance aligned meta-adaptation.
\newblock In: Proceedings of the IEEE/CVF International Conference on Computer
  Vision, pp. 8484--8494 (2021)

\bibitem{trx2021}
Perrett, T., Masullo, A., Burghardt, T., Mirmehdi, M., Damen, D.:
  Temporal-relational crosstransformers for few-shot action recognition.
\newblock arXiv preprint arXiv:2101.06184  (2021)

\bibitem{learningzoom2018}
Recasens, A., Kellnhofer, P., Stent, S., Matusik, W., Torralba, A.: Learning to
  zoom: a saliency-based sampling layer for neural networks.
\newblock In: Proceedings of the European Conference on Computer Vision (ECCV),
  pp. 51--66 (2018)

\bibitem{prototypical2017}
Snell, J., Swersky, K., Zemel, R.S.: Prototypical networks for few-shot
  learning.
\newblock arXiv preprint arXiv:1703.05175  (2017)

\bibitem{ucf101}
Soomro, K., Zamir, A.R., Shah, M.: Ucf101: A dataset of 101 human actions
  classes from videos in the wild.
\newblock arXiv preprint arXiv:1212.0402  (2012)

\bibitem{relationNet2018}
Sung, F., Yang, Y., Zhang, L., Xiang, T., Torr, P.H., Hospedales, T.M.:
  Learning to compare: Relation network for few-shot learning.
\newblock In: Proceedings of the IEEE conference on computer vision and pattern
  recognition, pp. 1199--1208 (2018)

\bibitem{vsd}
Tian, X., Zhang, Z., Lin, S., Qu, Y., Xie, Y., Ma, L.: Farewell to mutual
  information: Variational distillation for cross-modal person
  re-identification.
\newblock In: Proceedings of the IEEE/CVF Conference on Computer Vision and
  Pattern Recognition, pp. 1522--1531 (2021)

\bibitem{c3d2015}
Tran, D., Bourdev, L., Fergus, R., Torresani, L., Paluri, M.: Learning
  spatiotemporal features with 3d convolutional networks.
\newblock In: Proceedings of the IEEE international conference on computer
  vision, pp. 4489--4497 (2015)

\bibitem{tschannen2019mutual}
Tschannen, M., Djolonga, J., Rubenstein, P.K., Gelly, S., Lucic, M.: On mutual
  information maximization for representation learning.
\newblock arXiv preprint arXiv:1907.13625  (2019)

\bibitem{transfomer2017}
Vaswani, A., Shazeer, N., Parmar, N., Uszkoreit, J., Jones, L., Gomez, A.N.,
  Kaiser, L., Polosukhin, I.: Attention is all you need.
\newblock arXiv preprint arXiv:1706.03762  (2017)

\bibitem{matching2016}
Vinyals, O., Blundell, C., Lillicrap, T., Kavukcuoglu, K., Wierstra, D.:
  Matching networks for one shot learning.
\newblock arXiv preprint arXiv:1606.04080  (2016)

\bibitem{TSN2016}
Wang, L., Xiong, Y., Wang, Z., Qiao, Y., Lin, D., Tang, X., Van~Gool, L.:
  Temporal segment networks: Towards good practices for deep action
  recognition.
\newblock In: European conference on computer vision, pp. 20--36. Springer
  (2016)

\bibitem{HyRSM2022}
Wang, X., Zhang, S., Qing, Z., Tang, M., Zuo, Z., Gao, C., Jin, R., Sang, N.:
  Hybrid relation guided set matching for few-shot action recognition.
\newblock In: Proceedings of the IEEE/CVF Conference on Computer Vision and
  Pattern Recognition, pp. 19948--19957 (2022)

\bibitem{adafocus2021}
Wang, Y., Chen, Z., Jiang, H., Song, S., Han, Y., Huang, G.: Adaptive focus for
  efficient video recognition.
\newblock In: Proceedings of the IEEE/CVF International Conference on Computer
  Vision, pp. 16249--16258 (2021)

\bibitem{crop1}
Wang, Y., Lv, K., Huang, R., Song, S., Yang, L., Huang, G.: Glance and focus: a
  dynamic approach to reducing spatial redundancy in image classification.
\newblock Advances in Neural Information Processing Systems \textbf{33},
  2432--2444 (2020)

\bibitem{adafocusv2}
Wang, Y., Yue, Y., Lin, Y., Jiang, H., Lai, Z., Kulikov, V., Orlov, N., Shi,
  H., Huang, G.: Adafocus v2: End-to-end training of spatial dynamic networks
  for video recognition.
\newblock arXiv preprint arXiv:2112.14238  (2021)

\bibitem{fsl-augment1}
Wang, Y.X., Girshick, R., Hebert, M., Hariharan, B.: Low-shot learning from
  imaginary data.
\newblock In: Proceedings of the IEEE conference on computer vision and pattern
  recognition, pp. 7278--7286 (2018)

\bibitem{marl2019}
Wu, W., He, D., Tan, X., Chen, S., Wen, S.: Multi-agent reinforcement learning
  based frame sampling for effective untrimmed video recognition.
\newblock In: Proceedings of the IEEE/CVF International Conference on Computer
  Vision, pp. 6222--6231 (2019)

\bibitem{adaframe2018}
Wu, Z., Xiong, C., Ma, C.Y., Socher, R., Davis, L.S.: Adaframe: Adaptive frame
  selection for fast video recognition.
\newblock In: Proceedings of the IEEE/CVF Conference on Computer Vision and
  Pattern Recognition, pp. 1278--1287 (2019)

\bibitem{ARN2020}
Zhang, H., Zhang, L., Qi, X., Li, H., Torr, P.H., Koniusz, P.: Few-shot action
  recognition with permutation-invariant attention.
\newblock In: Proceedings of the European Conference on Computer Vision (ECCV).
  Springer (2020)

\bibitem{itanet}
Zhang, S., Zhou, J., He, X.: Learning implicit temporal alignment for few-shot
  video classification.
\newblock arXiv preprint arXiv:2105.04823  (2021)

\bibitem{yufeng}
Zhang, Y., Ding, L., Li, Y., Lin, W., Zhao, M., Yu, X., Zhan, Y.: A regional
  distance regression network for monocular object distance estimation.
\newblock Journal of Visual Communication and Image Representation \textbf{79},
  103224 (2021)

\bibitem{trilinear2019}
Zheng, H., Fu, J., Zha, Z.J., Luo, J.: Looking for the devil in the details:
  Learning trilinear attention sampling network for fine-grained image
  recognition.
\newblock In: Proceedings of the IEEE/CVF Conference on Computer Vision and
  Pattern Recognition, pp. 5012--5021 (2019)

\bibitem{HCL}
Zheng, S., Chen, S., Jin, Q.: Few-shot action recognition with hierarchical
  matching and contrastive learning.
\newblock In: Computer Vision--ECCV 2022: 17th European Conference, Tel Aviv,
  Israel, October 23--27, 2022, Proceedings, Part IV, pp. 297--313. Springer
  (2022)

\bibitem{mgsampler}
Zhi, Y., Tong, Z., Wang, L., Wu, G.: Mgsampler: An explainable sampling
  strategy for video action recognition.
\newblock In: Proceedings of the IEEE/CVF International Conference on Computer
  Vision, pp. 1513--1522 (2021)

\bibitem{compound2018}
Zhu, L., Yang, Y.: Compound memory networks for few-shot video classification.
\newblock In: Proceedings of the European Conference on Computer Vision (ECCV),
  pp. 751--766 (2018)

\bibitem{cmn-j2020}
Zhu, L., Yang, Y.: Label independent memory for semi-supervised few-shot video
  classification.
\newblock IEEE Transactions on Pattern Analysis and Machine Intelligence
  \textbf{44}(1), 273--285 (2020)

\end{thebibliography}

	
\end{document}